\documentclass[10pt,twocolumn,letterpaper]{article}

\usepackage[pagenumbers]{cvpr} 

\definecolor{cvprblue}{rgb}{0.21,0.49,0.74}
\usepackage[pagebackref,breaklinks,colorlinks,allcolors=cvprblue]{hyperref}

\usepackage{url}            
\usepackage{booktabs}       
\usepackage{amsfonts}       
\usepackage{nicefrac}       
\usepackage{microtype}      
\usepackage{algorithm}
\usepackage{graphicx}
\usepackage{makecell}
\usepackage{multirow}
\usepackage{bm}
\usepackage{booktabs} 
\usepackage{overpic}
\usepackage{pict2e}
\usepackage{amsmath}
\usepackage{colortbl}
\usepackage{rotating}
\usepackage{amssymb}
\usepackage{pifont}
\usepackage{subcaption}
\usepackage{listings}
\usepackage{setspace}
\usepackage[capitalize]{cleveref}
\crefname{section}{Sec.}{Secs.}
\Crefname{section}{Section}{Sections}
\Crefname{table}{Table}{Tables}
\crefname{table}{Tab.}{Tabs.}

\definecolor{myblue}{RGB}{169,196,235}
\definecolor{mygreen}{RGB}{213,232,212}
\definecolor{mygray}{RGB}{191,191,191}
\definecolor{mycolor}{RGB}{184,96,41}

\ifdefined \GramaCheck
\newcommand{\CheckRmv}[1]{}
\newcommand{\figref}[1]{Figure 1}%
\newcommand{\tabref}[1]{Table 1}%
\newcommand{\secref}[1]{Section 1}
\renewcommand{\eqref}[1]{Equation 1}
\else
\newcommand{\CheckRmv}[1]{#1}
\newcommand{\figref}[1]{Fig.~\ref{#1}}%
\newcommand{\tabref}[1]{Tab.~\ref{#1}}%
\newcommand{\secref}[1]{Sec.~\ref{#1}}

\renewcommand{\eqref}[1]{Eqn.~(\ref{#1})}
\fi

\def\ie{\emph{i.e.,~}}
\def\eg{\emph{e.g.,~}}

\def\paperID{3124} 
\def\confName{CVPR}
\def\confYear{2025}

\title{\emph{v}-CLR: View-Consistent Learning for Open-World Instance Segmentation}

\author{
Chang-Bin Zhang$^{1}$ \qquad\qquad
Jinhong Ni$^{1}$ \qquad\qquad
Yujie Zhong$^{2}$ \qquad\qquad
Kai Han$^{1}\thanks{Corresponding author.}$
\\
$^1$Visual AI Lab, The University of Hong Kong\qquad
$^2$Meituan Inc.\\
\texttt{\small \{cbzhang, jhni\}@connect.hku.hk} \qquad
\texttt{\small jaszhong@hotmail.com} \qquad
\texttt{\small kaihanx@hku.hk}
}

\begin{document}
\maketitle

\begin{abstract}
In this paper, we address the challenging problem of open-world instance segmentation.
  Existing works have shown that vanilla visual networks are biased toward learning appearance information, \eg texture, to recognize objects. This implicit bias causes the model to fail in detecting novel objects with unseen textures in the open-world setting. 
  To address this challenge, we propose a learning framework, called view-Consistent LeaRning (v-CLR), which aims to enforce the model to learn appearance-invariant representations for robust instance segmentation. 
  In v-CLR, we first introduce additional views for each image, where the texture undergoes significant alterations while preserving the image's underlying structure. 
  We then encourage the model to learn the appearance-invariant representation by enforcing the consistency between object features across different views, for which we obtain class-agnostic object proposals using off-the-shelf unsupervised models that possess strong object-awareness. These proposals enable cross-view object feature matching, greatly reducing the appearance dependency while enhancing the object-awareness. 
  We thoroughly evaluate our method on public benchmarks under both cross-class and cross-dataset settings, achieving state-of-the-art performance. 
  Project page: \url{https://visual-ai.github.io/vclr}
\end{abstract}

\section{Introduction}
\label{sec:intro}
Modern object detectors~\cite{fastrcnn,detr,deformabledetr,zhang2024mr} and instance segmentors~\cite{maskrcnn,maskformer,maskdino} have achieved many milestones.
However, these detectors are based on the assumption of pre-defined taxonomy classes.
Despite recent open-vocabulary detectors~\cite{gu2021open,kim2023region} can be extended to larger taxonomy classes benefiting from the foundation model pre-trained on large-scale text-image pairs, these models are still limited by the finite taxonomy classes in the pre-trained data.
In some realistic applications, models are required to identify out-of-taxonomy classes.
Thus, recognizing objects in the open world has been increasingly interesting and challenging.

In open-world instance segmentation, models are trained on a set of predefined known classes and are evaluated to localize unknown objects during inference.
Following~\cite{oln,udos,ldet,sword}, we regard the open-world instance segmentor as a class-agnostic object discovery model.
One straightforward solution is to train a class-agnostic detector on labeled instances of known classes, \ie performing binary object detection given ground truth labels from the known classes, and hope the models capture transferable features that generalize to unknown objects.
However, various studies~\cite{brendel2019approximating,gatys2017texture,ballester2016performance,geirhos2018imagenet} have demonstrated that neural networks exhibit a preference to capture texture information when recognizing objects. This hinders the model's ability to generalize in the open-world setting, especially to unknown objects with unseen textures.

\begin{figure}[!tp] 
  \centering
  \small
  \begin{overpic}[width=.99\linewidth]{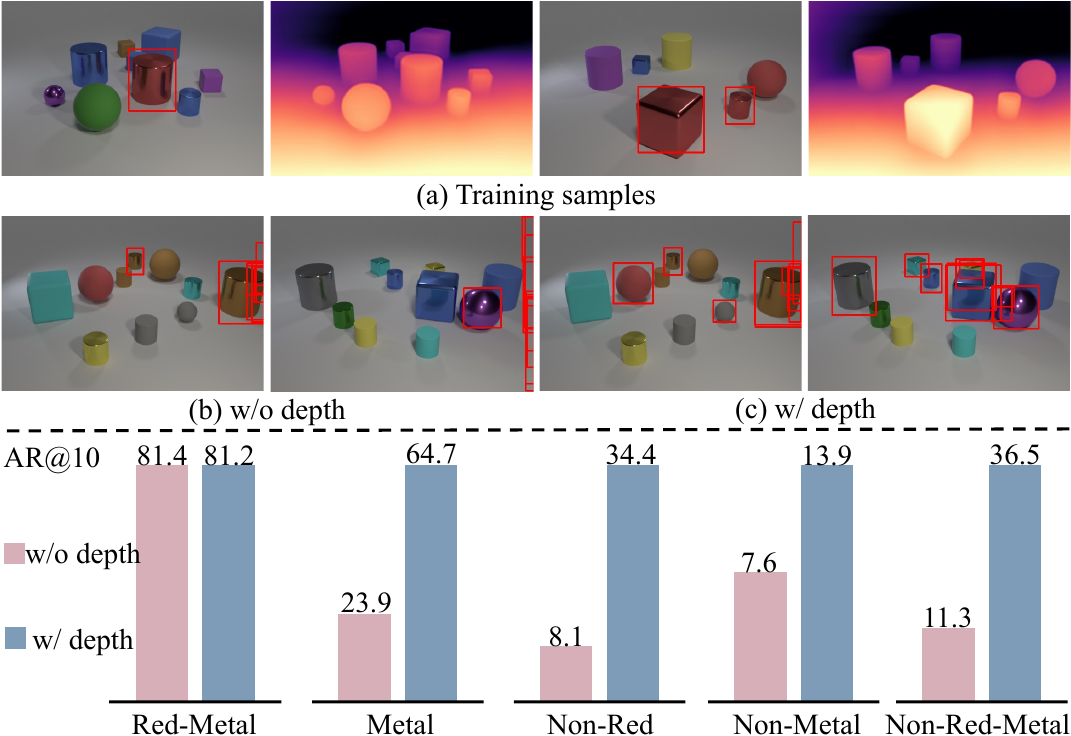}

  \end{overpic}
    \vskip 0.1in
  \caption{\textbf{Toy example on the CLEVR~\cite{clevr} dataset.} The model regards \textit{red-metal} objects as the known class and is evaluated on different subsets in terms of AR@10.
  We train the model with and without incorporating depth image data, respectively. The prediction results are displayed in the middle row.
  }\label{fig:teaser}
\end{figure}

To motivate the necessity of capitalizing appearance-invariant information, we showcase a toy open-world example run on the CLEVR~\cite{clevr} dataset in~\figref{fig:teaser}.
In this example, we treat the \emph{red metal} objects as the known class, and evaluate the model on detecting various other types of objects (with other colors or materials).
We show an example of training samples in~\figref{fig:teaser} (a), where each sample consists of a natural image and a colorized depth map.
We label the known class, \ie the red metal objects, with red bounding boxes.
We then train a vanilla detector as the baseline model using only natural images as input, and a model incorporating colorized depth images.
The evaluation results on various object subsets involving different colors and materials in~\figref{fig:teaser} demonstrate that the model trained with depth images exhibits a much better generalization to novel objects.
This toy example verifies the problem that the vanilla baseline models suffer from poor generalization due to the appearance bias, and emphasizes the importance of including appearance-invariant information to guide representation learning.

To overcome this challenge, we propose a \emph{view-Consistent LeaRning} framework, dubbed \emph{v}-CLR, to encourage the model to learn appearance-invariant representation for novel object discovery.
To achieve this, we first transform images into multiple appearance-invariant views, from which we propose a feature-matching objective to enforce cross-view feature consistency.
This objective alone would be insufficient as there is no guarantee that similar features correspond to objects, we thus adopt off-the-shelf general object proposals to ensure optimized representations are object-oriented.
Specifically, we first exploit the appearance-invariant information by transforming the natural images into various other domains, \eg colorized depth images.
Intuitively speaking, these transformations destroy or overwrite the appearance information from the natural image domain while preserving the original structures, thus encouraging the model to capitalize information other than appearance.

To facilitate appearance-invariant representation learning and effectively utilize training data containing multiple views, we build on top of DETR-like architectures~\cite{detr,deformabledetr,dino}, in which we enforce representation consistency across different views of the same image by matching similar queries.
By doing this, we naturally circumvent the problem of implicit appearance bias by empowering the model to capture consistent cross-view information.
However, naively enforcing such consistency may still fail in reality.
The reason is that even if the model extracts similar representations across different views, it does not necessarily imply these representations are object-related. 
To sidestep this problem, we adopt pre-trained unsupervised instance detectors, \eg CutLER~\cite{cutler}, to generate object proposals. 
These off-the-shelf instance detectors exhibit high instance awareness, for which we explicitly match the queries from different views with the object proposals to ensure these paired queries are object-oriented.
To this end, we have devised a learning framework to allow models to capture object-related consistent appearance-invariant representations, enabling transferability to novel objects in open-world scenarios.

We conduct extensive experiments on various benchmarks, including COCO 2017~\cite{coco}, LVIS~\cite{lvis}, UVO~\cite{uvo}, and Objects365~\cite{objects365}, under cross-categories and cross-datasets settings. Our proposed learning framework consistently achieves state-of-the-art performance on several benchmarks in the open-world setting.

\section{Related Work} 
\noindent\textbf{Object Detection and Instance Segmentation.}
DETR~\cite{detr} and its follow-up works~\cite{deformabledetr,dino,meng2021conditional,liu2022dab,li2022dn,gao2022adamixer,zhang2025humanoidpano} achieve an end-to-end detector with remarkable performance, improving transformer architecture~\cite{deformabledetr, gao2022adamixer,meng2021conditional,liu2022dab}, training efficiency~\cite{li2022dn,dino,chen2022group,jia2023detrs,hu2024dac,zhang2024mr} and label assignment~\cite{liu2023detection,cai2023align,teng2023stageinteractor}.
MaskDINO~\cite{maskdino} develop a unified model for object detection and instance segmentation. 
Benefiting from the powerful self-supervised learning~\cite{dinov1,oquab2023dinov2,mae,wang2021dense}, unsupervised instance segmentation~\cite{wang2022freesolo,wang2023tokencut} has received increasing interest by discovering pixel-level pseudo annotations automatically.
Thanks to strong object-awareness from self-supervised pretrained models~\cite{oquab2023dinov2}, CutLER~\cite{cutler} constructs a large-scale training set with pseudo masks, \eg ImageNet dataset, and train an instance segmentation model without any human annotation.
In our work, we utilize the CutLER pre-trained on the ImageNet as a general objects proposal network.

\noindent\textbf{Open-world Instance Segmentation.}
To promote the applications of modern object detectors in realistic scenarios, recent arts~\cite{oln,ggn} propose open-world instance segmentation.
To avoid suppressing potential unknown objects in background regions, OLN~\cite{oln} replaced the classification branch in Mask-RCNN~\cite{maskrcnn} with a localization-aware score.
LDET~\cite{ldet} proposed to synthesize training images by combining labeled objects and predefined background texture by copy-paste~\cite{copypaste}.
Segprompt~\cite{segprompt} utilizes prompting designation to segment novel objects.
Some other methods~\cite{ggn,udos,good} design variant mechanisms to discover potential unknown objects in training images, including grouping pixels~\cite{ggn}, leveraging prior mask~\cite{udos} by MCG~\cite{mcg} and imposing geometry information~\cite{good}.
SWORD~\cite{sword} explores applying DETR-based model~\cite{deformabledetr} on the open-world instance segmentation. 
SOS~\cite{SOS} propose to discover potential unlabeled objects by SAM~\cite{Kirillov2023SegmentA} with DINOv2~\cite{oquab2023dinov2} activation point as prompt.
In our work, we conduct experiments based on the Deformable-DETR~\cite{deformabledetr} and DINO-DETR~\cite{dino}, respectively.

\noindent\textbf{Texture-Invariant Representations.}
Within the domain of generalization and adaptation, models are designed to utilize source domain training data to achieve effective performance on a different target domain, assuming that both domains share the same set of semantic categories.
To successfully adapt to the target domain, which may exhibit different styles from the source, current approaches~\cite{Choi2021RobustNetID,Lee2022WildNetLD,Wu2022SingleDomainGO,Huang2023StylePC,Kim2023TextureLD,yang2020fda,kim2020learning} incorporate style transfer techniques to modify training images to either the target or an arbitrary style.
Our approach emphasizes the transfer of knowledge across different semantic classes instead of across different domains. 
The challenges between domain shifts and semantic shifts are orthogonal~\cite{wang2024dissect} and the techniques for domain shifts are not suitable for semantic shifts. 
To this end, our method can leverage any transformation views with unified features among objects, extending beyond mere style transfer. 
For instance, depth images and edge maps can also be utilized to develop unified object representations, which are typically unsuitable for domain generalization due to their lack of class discriminability.
In our approach, depth images serve as the primary transformation view, while stylized images and edge maps can be optionally used as auxiliary views.

\begin{figure*}[!tp] 
  \centering
  \small
  \begin{overpic}[width=.99\linewidth]{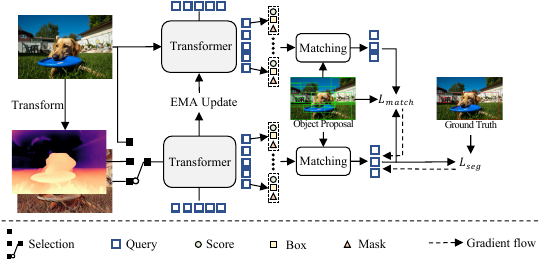} 
  \end{overpic}
  \caption{\textbf{Illustration of \emph{v}-CLR.} Our learning framework consists of two branches, the natural image branch (top) and the transformed image branch (bottom). Both branches adopt transformers to make predictions, which are then matched with the object proposals to obtain optimized object queries. We compute a matching loss $L_{match}$ which enforces the matched object-oriented query pairs from the two branches to be similar. We finally compute the ordinary segmentation loss $L_{gt}$ using the ground truth labels. The transformer in the natural image branch is updated as an EMA model of the transformed image branch.
  }\label{fig:method}
\end{figure*}

\section{Method}\label{sec:method}

\noindent\textbf{Problem Statement.}
Open-world instance segmentation aims to localize as many novel objects as possible during test time.
Formally, the training labels are first divided into two sets of known classes ($\mathcal{C}_{base}$) and unknown classes ($\mathcal{C}_{novel}$), with no overlap between them (\ie $\mathcal{C}_{base}\cap\mathcal{C}_{novel}=\emptyset$).
For each training sample image $I$ and its associated set of annotations $C$, we train the models only on the annotations of known classes, in a class-agnostic manner .
During test time, we evaluate the model's capability of generalizing on the set of unknown classes ($\mathcal{C}_{novel}$).

\subsection{Method Overview}\label{sec:method-overview}
\noindent\textbf{Architecture.}
Inspired by instance segmentation models with transformer~\cite{maskformer,maskdino}, we decorate the Deformable-DETR~\cite{deformabledetr} and DINO-DETR~\cite{dino} into the instance segmentation model.
Specifically, following~\cite{maskformer, maskdino}, each query predicts a prototype for a corresponding instance, and then the model will predict the instance segmentation map by computing the similarity between the output prototype and the pyramid features of the transformer encoder.

\noindent\textbf{Appearance-Invariant Transformation.}
To enable such an appearance-invariant representation learning, we first leverage off-the-shelf image transformation to \emph{overwrite the appearance from the natural images while leaving the overall structural contents intact}. 
The intuition is that we circumvent the texture bias~\cite{brendel2019approximating,gatys2017texture,ballester2016performance,geirhos2018imagenet} by allowing the model to learn consistent and transferable representations from different image transformations. 
We adopt colorized depth maps~\cite{bhat2023zoedepth} as the major transformation in this work, and with an additional auxiliary transformation \eg art-stylizing~\cite{wu2021styleformer} and edge map~\cite{Xie2015HolisticallyNestedED}, while we highlight that our method is not strictly bound by any transformation method so long as they suffice the aforementioned criteria. 
Complementing the two transformations with the natural images gives us three views, \ie natural images, colorized depth maps, and one additional auxiliary view, for each training sample, from which we randomly select one view per sample with equal probability during training. 
To further destroy the appearance of objects, we apply random cropping and resizing to an image patch, subsequently integrating it with the original image.
These various views play a crucial role in our method as described in the following section.

\subsection{Appearance-Invariant Representation}\label{sec:VCR}
Existing works have shown evidence that neural networks are biased toward learning appearance information, \eg texture, to differentiate different objects~\cite{brendel2019approximating,gatys2017texture,ballester2016performance,geirhos2018imagenet}. This tendency of relying on appearance information inhibits the generalization ability to novel classes especially when unseen textures are presented during inference. To overcome this challenge, we devise a learning framework so that the model learns appearance-invariant representations complementing the appearance information and, thus are generalizable and unbiased during inference. Our proposed method is detailed below. Roughly speaking, the key to this learning framework is to enforce representation consistency by maximizing the query feature similarity between the transformed views and the natural image.

Our learning framework comprises two branches: the natural image branch, which always receives natural images as inputs; and the transformed image branch, which randomly processes any of the transformed images or the original natural image with equal probability. 
Both branches then utilize the adapted DETR transformer architectures~\cite{detr,deformabledetr,dino} to make sets of predictions, where each prediction consists of a classification score, a predicted bounding box, and a predicted segmentation mask. We refer the readers to the \emph{Model Architecture} paragraph in~\secref{sec:method-overview} for details regarding how we adopt detection transformers for instance segmentation. Following existing self-supervised learning frameworks~\cite{mocov1,mocov2,simsiam,siamesedetr}, to prevent feature collapsing, we update the transformer in the natural image branch as an exponential moving average (EMA) model of the transformed image branch.

\noindent\textbf{Object-centric Learning by Object Proposals.}
At first glance, it seems to be feasible at this stage to ensure representation consistency on the query features outputted from the two branches. However, a high similarity between the matched queries does not necessarily imply the model learning informative representation. An example is when models capture shortcut solutions where the extracted representations are irrelevant to the objects. 
In the context of open-world learning, a lack of correlation with the objects can cause failure in generalization. 
Thanks to the high instance awareness of the large-scale pre-trained instance detectors~\cite{cutler}, we sidestep the problem of the model falling into object-irrelevant solutions by leveraging these pre-trained detectors to provide object proposals.
These object proposals serve as a medium to match object-related queries from both branches, thus ensuring our learning framework can learn meaningful object-oriented representation to be successfully transferred to open-world settings.

\noindent\textbf{View-Consistent Learning.}
Given the multiple transformed views of an image, we hope a model can learn to extract consistent characteristics shared across different views of the same image. To facilitate such training, we propose \emph{view-Consistent LeaRning}. An overview of our method is illustrated in~\figref{fig:method}.

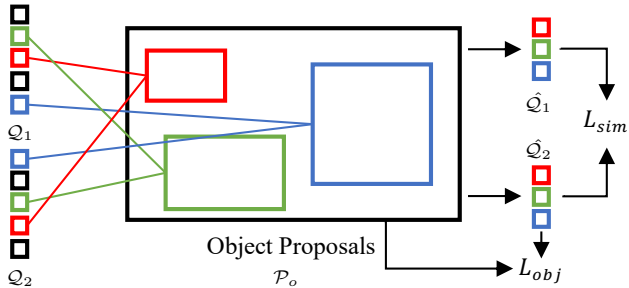
\begin{figure}[t]
\begin{overpic}[width=.99\linewidth]{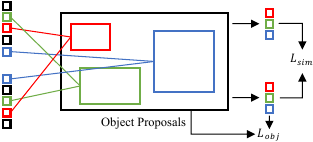} 
    \put(0, 0) {\scriptsize $\mathcal{Q}_2$}
    \put(0, 23.5) {\scriptsize $\mathcal{Q}_1$}
    \put(83.5, 28) {\scriptsize $\hat{\mathcal{Q}_1}$}
    \put(83.5, 20.5) {\scriptsize $\hat{\mathcal{Q}_2}$}
    \put(43, 0) {\scriptsize $\mathcal{P}_o$}
  \end{overpic}
\vspace{2mm}
\caption{\textbf{Illustration of object feature matching in \emph{v}-CLR.} Let $\mathcal{Q}_1$ and $\mathcal{Q}_2$ represent the query outputs from the EMA teacher model and the student model, respectively. Predictions associated with object proposals demonstrating poor localization quality are removed, resulting in paired $\hat{\mathcal{Q}_1}$ and $\hat{\mathcal{Q}_2}$, and the objective $L_{sim}$ is utilized to maximize feature similarity between each pair. Concurrently, the student model is trained using these object proposals.
}\label{fig:lossmatching}
\end{figure}

\noindent\textbf{Object Feature Matching.}
We introduce the object feature matching in our view-consistent learning pipeline in detail. The overall illustration of the matching objective is shown in~\figref{fig:lossmatching}. 
Formally, denote the sets of predictions from two branches as $\mathcal{P}_1$ and $\mathcal{P}_2$, and the set of extracted object proposals as $\mathcal{P}_o$, where each set $\mathcal{P}=\{(\hat{p_i},\hat{b_i},\hat{m_i})\}$ consists of tuples of class score $\hat{p_i}$, bounding box $\hat{b_i}$, and segmentation mask $\hat{m_i}$, for $i=1,\dots,|\mathcal{P}|$. We also have the sets of queries $\mathcal{Q}_1$ and $\mathcal{Q}_2$ associated with the prediction sets, where we have $|\mathcal{Q}_i|=|\mathcal{P}_i|$ for $i=1,2$. Following the previous works~\cite{detr,kuhn1955hungarian}, for each proposal in $\mathcal{P}_o$, we find the optimal sets $\Hat{\mathcal{P}_1}$ and $\Hat{\mathcal{P}_2}$ for the two sets of predictions by minimizing the matching cost. The sets $\mathcal{P}_o$, $\Hat{\mathcal{P}_1}$, and $\Hat{\mathcal{P}_2}$ forms $\Tilde{N}$ one-to-one triplets. 

\noindent\textbf{Training Objectives.}
We denote the optimal sets of queries as $\Hat{\mathcal{Q}_1}$ and $\Hat{\mathcal{Q}_2}$ corresponding to the sets of predictions $\Hat{\mathcal{P}_1}$ and $\Hat{\mathcal{P}_2}$, for which we compute the cosine similarity matching loss:
\begin{equation*}
    L_{sim}=\frac{1}{\Tilde{N}}\sum_{q_i\in\Hat{\mathcal{Q}_i}}\left(1-\cos\left(q_1, q_2\right)\right),
\end{equation*}
where $\cos(q_1,q_2)$ denotes the cosine similarity between $q_1$ and $q_2$. Since we assume the object proposals to be reliably object-related, this may give us additional information for supervising the predicted boxes and segmentation maps. We thus compute the standard segmentation loss using the object proposals $L_{obj}$:
\begin{equation}\label{eq:Lobj}
\begin{aligned}
    L_{obj} = \lambda_{1}L_{dice} &+ \lambda_{2}L_{mask}+\\  &\lambda_{3}L_{score} +
    \lambda_{4}L_{box} +
    \lambda_{5}L_{giou},
\end{aligned}
\end{equation}
where $\lambda_i$ from now on denotes the loss weight factor. The total matching objective is computed as:
\begin{equation*}
    L_{match}=\lambda_{obj}L_{obj}+\lambda_{sim}L_{sim}.
\end{equation*}
The matching objective ensures the queries capture object-oriented appearance-invariant representations. We proceed to the regular segmentation loss using the ground truth labels. Formally, given the set of optimized transformed image queries $\hat{\mathcal{Q}_2}$ and the set of ground truth $\mathcal{G}$, we compute similar segmentation objective $L_{gt}$ as~\eqref{eq:Lobj} by replacing the object proposals $\mathcal{P}_o$ with $\mathcal{G}$.
The total training objective is then:
\begin{equation*}
    L=\lambda_{match}L_{match}+\lambda_{gt}L_{gt}.
\end{equation*}

\section{Experiments}\label{sec:exps}

\subsection{Setup}\label{sec:setup}
\noindent\textbf{Datasets and Evaluations.}
We conduct experiments in two popular open-world settings, cross-categories and cross-datasets, on the CLEVR~\cite{clevr}, COCO 2017~\cite{coco}, LVIS~\cite{lvis}, UVO~\cite{uvo} and Objects365~\cite{objects365} datasets.
The prior setting divides the object classes into known and unknown classes, whereas the latter setting tests the generalization ability of the model on another dataset containing unseen object classes.
Since the labels in validation images can not cover all objects, we apply the average recall (AR) over multiple IoU thresholds $[0.5,0.95]$ to measure the model's performance, while ignoring the average precision (AP) as previous arts~\cite{oln,ggn,sword}.
Following~\cite{sword,oln,udos}, \emph{the most widely concerned metric in this task is AR@100}, which is denoted by AR$_{100}$ in our paper.
As standard evaluation metrics on COCO, we use AR$^b$ and AR$^m$ to denote the results for predicted boxes and instance masks, respectively.
We additionally report the performance for small, medium, and large objects, denoted by AR$_{s/m/l}$ respectively.

\noindent\textbf{Implementation Details.}\label{sec:implement-detail}
We regard the model as a class-agnostic object detector in all experiments.
We apply the DINO-DETR~\cite{dino} with ResNet-50~\cite{he2016deep} as the backbone to perform instance segmentation. 
We adopt the common settings in DETR-like models~\cite{detr,dino,zhang2024mr}, \eg there are six layers in the transformer encoder and decoder, respectively.
We set the number of denoising queries~\cite{li2022dn} as 300.
Inspired by~\cite{maskformer,sword,maskdino}, we decorate the DINO-DETR with dynamic convolution for instance segmentation prediction.
Following~\cite{sword}, we use 1500 and 1000 queries in the transformer decoder when training on VOC and COCO classes, respectively.
We train the model for 8 epochs and the learning rate is decayed at the 7th epoch, while keeping other settings in the training schedule as fully-supervised object detectors.
In our experiments, $\lambda_{sim}$, $\lambda_{obj}$ and $\lambda_{gt}$ is set to 1, and coefficients in~\eqref{eq:Lobj} are the same as DINO~\cite{dino}.
We use the pre-trained Cascade-Mask-RCNN~\cite{cai2018cascade} as the object proposal network without any fine-tuning, which is trained by CutLER~\cite{cutler} with ResNet-50 as the backbone.

\subsection{Main Results}\label{sec:mainres}
To validate the effectiveness of our method, we conduct experiments in popular settings, including VOC $\rightarrow$ Non-VOC, COCO $\rightarrow$ LVIS, VOC $\rightarrow$ UVO, and COCO $\rightarrow$ Objects365, 
where $\mathcal{D}_A \rightarrow \mathcal{D}_B$ denotes training the model on dataset $\mathcal{D}_A$ and evaluating the transferability on the dataset $\mathcal{D}_B$.

\begin{table}[t]
    \centering
    \begin{tabular}{l|cccc}
         \toprule
        Method & AR$^b_{10}$ & AR$^b_{100}$ & AR$^m_{10}$ & AR$^m_{100}$ \\
        \midrule
        Mask-RCNN~\cite{maskrcnn}  & 10.2 & 23.5 & 7.9 & 17.7\\
        CutLER~\cite{cutler} & 19.9 & 34.5 & - & - \\
        OLN~\cite{oln}  & 18.0 & 33.5 & 16.9 & -\\
        LDET~\cite{ldet}  & 18.2 & 30.8 & 16.3 & 27.4\\
        GGN~\cite{ggn}  & 17.3 & 31.6 & 16.1 & 28.7\\
        GGN + OLN~\cite{oln} & 17.1 & 37.2 & 16.4 & 33.7 \\
        UDOS~\cite{udos} & - & 33.5 & - & 31.6 \\
        GOOD$^{\dag}$~\cite{good} & - & 39.3 & - & - \\
        \midrule
        Def-DETR~\cite{deformabledetr}& 12.2 & 27.4 & 10.2 & 22.7\\
        SWORD~\cite{sword} & 17.8 & 35.3 & 15.7 & 30.2\\
        \rowcolor{gray!20}\emph{v}-CLR (Def-DETR) & \underline{22.2} & \underline{40.3} & \underline{19.6} & \underline{33.7} \\
        \midrule
        DINO-DETR~\cite{dino} & 13.2 & 31.1 & 9.7 & 22.0 \\
        \rowcolor{gray!20}\emph{v}-CLR (DINO) & \textbf{22.5}  & \textbf{40.9} & \textbf{19.9} & \textbf{34.1}\\
        \bottomrule
    \end{tabular}
    \caption{\textbf{Evaluation results for novel classes in the VOC $\rightarrow$ Non-VOC setting.} The ${\dag}$ denotes the model is trained with bounding boxes only.} \label{tab:voc2nonvoc}
\end{table}

\begin{table}[t]
    \centering
    \begin{tabular}{l|cccc}
    \toprule
    Method & AR$^b_{10}$ & AR$^b_{100}$ & AR$^m_{10}$ & AR$^m_{100}$ \\
    \midrule
    Mask-RCNN~\cite{maskrcnn}  & 11.4 & 16.2 & 7.6 & 11.4\\
    LDET~\cite{ldet} & 16.0 & 31.9 & 12.3 & 25.2 \\
    \midrule
    Def-DETR~\cite{deformabledetr} & 13.5 & 33.5 & 9.5 & 25.3 \\
    SWORD~\cite{sword} & 16.8 & 43.1 & 13.3 & \underline{34.9} \\
    \rowcolor{gray!20}\emph{v}-CLR (Def-DETR) & \underline{20.3} & \underline{45.8} & \underline{16.1} & 34.6 \\
    \midrule
    DINO-DETR~\cite{dino} & 14.7 & 36.5 & 10.7 & 27.7 \\
    \rowcolor{gray!20}\emph{v}-CLR (DINO) & \textbf{21.0} & \textbf{47.2} & \textbf{16.8} & \textbf{35.9} \\
\bottomrule
    \end{tabular}
    \caption{\textbf{Evaluation results for novel classes in the VOC$\rightarrow$UVO setting.}}
    \label{tab:voc2uvo}
\end{table}

\noindent\textbf{\textbf{VOC $\rightarrow$ Non-VOC}.}
The VOC~\cite{voc} dataset includes 20 common classes in natural images, for which we train the model on VOC classes to verify the generalization capability of our method.
Specifically, the model is trained on the COCO 2017 training set with 20 VOC class labels, and tested on the other 60 Non-VOC classes on the COCO validation set. 
Following recent arts~\cite{sword,oln}, we also regard the prediction as a class-agnostic scheme, thus the most concerned evaluation metric is average recall (AR), especially AR@100.
As shown in~\tabref{tab:voc2nonvoc}, we report the AR@10 and AR@100 on the Non-VOC classes, respectively.
SOWRD~\cite{sword} firstly explore adapting DETR-based detector to discover novel objects, and propose some techniques based on popular Deformable-DETR~\cite{deformabledetr}, including stop-gradient, IoU-based branch, and one-to-many assignment.
However, we empirically find that vanilla DINO-DETR can achieve surprisingly strong performance with the help of denoising queries to accelerate training.
Therefore, we conduct experiments based on Deformable-DETR~\cite{deformabledetr} and DINO-DETR~\cite{dino} for a fair comparison, respectively.
Experimental results demonstrate that our method achieves state-of-the-art performance on all evaluation metrics in this setting.

\noindent\textbf{VOC $\rightarrow$ UVO.}
The UVO dataset~\cite{uvo} is a large-scale dataset designed for open-world segmentation, covering many kinds of objects in the wild.
To validate the cross-dataset generalization, we follow previous work to conduct experiments on the UVO dataset~\cite{ldet,sword}.
Specifically, the model is trained on the 20 VOC classes of COCO 2017 training set, and is evaluated on the UVO dense v1.0 validation set.
This split provides the category names of each instance, which allows us to split the novel classes and evaluate our model.
We report the experimental results in~\tabref{tab:voc2uvo}.
Compared with the previous state-of-the-art method, our method achieve a remarkable improvement of 2.7\% in terms of AR$_{100}^b$ based on Deformable-DETR~\cite{deformabledetr}.
We argue that the baseline model tends to suffer from the bias on the limited appearance of known classes. Benefiting from learning appearance-invariant information, our method improves more than 10\% both on AR$_{100}^b$ and AR$_{100}^m$.

\noindent\textbf{COCO $\rightarrow$ LVIS.}
The LVIS dataset~\cite{lvis} enlarges the taxonomy of COCO, containing more than 1200 classes where a large number of classes are disjoint with COCO classes.
In this setting, to verify the generalization ability on larger known taxonomy, the model is trained on 80 classes of COCO 2017 training set, and evaluated on other disjoint classes in LVIS validation set.
As shown in~\tabref{tab:coco2lvis}, although vanilla DINO-DETR reaches a better performance than SWORD~\cite{sword}, our method can additionally improve the baseline by 3.2\% in terms of AR$_{100}^b$.
Our proposed method outperforms SWORD by about 3.7\% and 1.9\% in terms of AR$_{100}^b$ and AR$_{100}^m$.
We argue this improvement arises from our proposed training framework, which encourages the model to learn appearance-invariant cues to discover potential objects.

\begin{table}[t]
    \centering
    \begin{tabular}{l|cccc}
    \toprule
    Method & AR$^b_{10}$ & AR$^b_{100}$ & AR$^m_{10}$ & AR$^m_{100}$ \\
    \midrule
    Mask-RCNN~\cite{maskrcnn}  & 6.1 & 19.4 & 5.6 & 17.2\\
    GGN~\cite{ggn}  & 7.6 & 22.4 & 7.2 & 20.4\\
    \midrule
    Def-DETR~\cite{deformabledetr}& 6.3 & 19.4 & 5.5 & 16.4\\
    SWORD~\cite{sword} & \underline{8.8} & 23.5 & \textbf{8.0} & 20.4 \\
    \rowcolor{gray!20}\emph{v}-CLR (Def-DETR) & \textbf{9.4} & \underline{27.2} & \textbf{8.0} & \underline{22.3} \\
    \midrule
    DINO-DETR~\cite{dino} & 8.5 & 25.2 & 7.4 & 21.0 \\
    \rowcolor{gray!20}\emph{v}-CLR (DINO) & \underline{9.3} & \textbf{28.4} & \underline{7.9} & \textbf{23.6}\\
    \bottomrule
    \end{tabular}
    \caption{\textbf{Evaluation results for novel classes in the COCO $\rightarrow$ LVIS setting.}}
    \label{tab:coco2lvis}
\end{table}

\begin{table}[t]
    \centering
    \setlength{\tabcolsep}{4pt}
    \begin{tabular}{l|ccccc}
         \toprule 
        Method& AR$^b_{10}$ & AR$^b_{100}$ & AR$^b_s$ & AR$^b_m$ & AR$^b_l$ \\ 
        \midrule
        Mask-RCNN~\cite{maskrcnn}  & 19.3 & 32.8 & 18.2 & 36.4 & 43.5   \\
        LDET~\cite{ldet}  & \underline{20.0} & 36.8 & 20.7 & 40.5 & 48.9 \\
        \midrule
        Def-DETR~\cite{deformabledetr}  & 19.0 & 40.1 & 22.8 & 43.4 & 54.1 \\
        SWORD~\cite{sword}  & \textbf{22.8} & 43.9 & 25.0 & 48.6 & 57.6 \\
        \rowcolor{gray!20}\emph{v}-CLR (Def-DETR) & 19.4 & 45.9 & 23.8 & 49.3 & \underline{62.8} \\
        \midrule
        DINO-DETR~\cite{dino}  & 19.0 & \underline{46.4} & \textbf{28.8} & \underline{50.0} & 58.6 \\
        \rowcolor{gray!20}\emph{v}-CLR (DINO) & 19.7 & \textbf{47.9} & \underline{26.2} & \textbf{51.6} & \textbf{64.0}  \\
        \bottomrule
    \end{tabular}
    \caption{\textbf{Evaluation results for novel classes in the COCO $\rightarrow$ Objects365 setting.}}
    \label{tab:coco2objects365}
\end{table}

\noindent\textbf{COCO $\rightarrow$ Objects365.}
The Objects365 dataset~\cite{objects365} includes 365 common classes which is much larger than COCO taxonomy.
As shown in~\tabref{tab:coco2objects365}, the model is trained on COCO 80 classes and evaluated on the novel classes of Objects365.
Since this dataset does not provide the instance mask annotation, we only evaluate the performance of bounding box prediction.
The experimental results demonstrate that our method can outperform SWORD~\cite{sword} by 2\% in terms of AR$_{100}^b$.
We also report the performance of different methods on the small, medium, and large objects, respectively.
We observe that our method performs slightly worse on small objects than vanilla DINO-DETR and explore the potential reasons in~\secref{sec:ablationstudy}.

\begin{figure*}[t] 
  \centering
  \small
  \begin{overpic}[width=.99\linewidth,page=1]{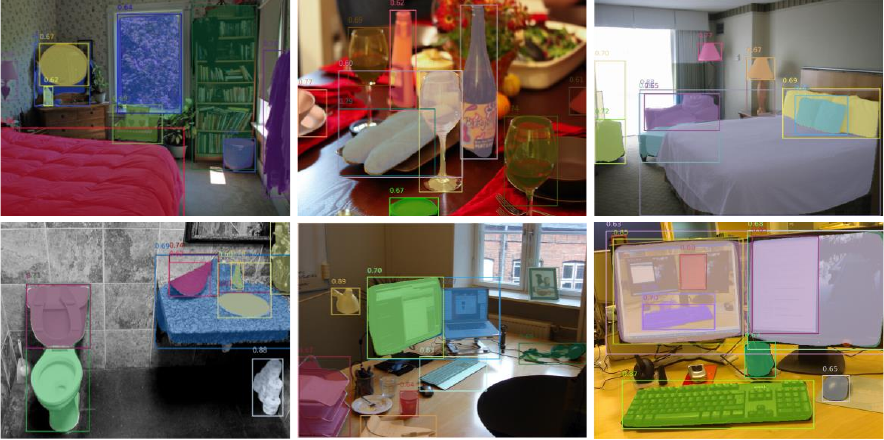} 
  \end{overpic}
  \vskip 0.1in
  \caption{\textbf{Qualitative results of our method on COCO 2017 validation set.} The model is trained on 20 VOC classes. We show the top-10 predicted instances according to the prediction confidence.
  }\label{fig:visualization}
\end{figure*}

\subsection{Qualitative Results}\label{sec:qualitativeres}
We visualize prediction results of our method on the COCO 2017 validation set in~\figref{fig:visualization}.
The model is trained on 20 VOC classes of the COCO 2017 training set.
For each image, we show the \emph{top}-10 predicted instances according to the prediction confidence.

\subsection{Ablation Study}\label{sec:ablationstudy}

\noindent\textbf{Ablation study of components.}
To validate the effectiveness of each component in our method, we conduct ablation studies in the VOC$\rightarrow$Non-VOC setting, as shown in~\tabref{tab:ablationcomponents}.
Initially, incorporating general object proposals results in a 6\% improvement over the vanilla DINO-DETR. Leveraging the colorized depth and auxiliary views introduced in our method, the detector achieves 40.0\% in terms of AR$_{100}^b$, marking a 2.3\% improvement over the strong baseline with $L_{obj}$ only. 
Based on this, our consistent constraint training objective yields an additional 0.2\% improvement, raising AR$_{100}^b$ to 40.2\%. To further enforce instance consistency, we filter paired object queries from the two branches before computing $L_{sim}$. This filtering results in a 0.7\% improvement in AR$_{100}^b$, culminating in our final model with an AR$_{100}^b$ of 40.9\%.
Without CutLER~\cite{cutler} proposals, our method reaches 30.7\% AR$^m_{100}$, achieving 8.6\% improvement over the baseline model.
Notably, general object proposals may be less effective when few or no unknown objects appear in the training images.
In~\tabref{tab:ablationcomponents}, following~\cite{good,udos}, all experiments, except for the baseline model (first row), also utilize the unlabeled images in the training set in the VOC$\rightarrow$Non-VOC setting.
When CutLER object proposals are not applied, we use the trained baseline model to provide annotations for these unlabeled images to ensure a fair comparison.

\begin{table}[t]
    \centering
    \small
    \setlength{\tabcolsep}{2pt}
    \begin{tabular}{ccccc|cc}
    \toprule
       $L_{gt}$ & $L_{obj}$ & Transform. & $L_{sim}$ & filtering & AR$_{100}^b$ & AR$_{100}^m$\\
    \midrule
     \ding{52}&&  & & & 31.1 & 22.0 \\
    \ding{52}&\ding{52} &  & & & 37.7 (\textcolor{blue}{+6.6}) & 31.2 (\textcolor{blue}{+9.2}) \\
    \ding{52}&\ding{52} & \ding{52} & & & 40.0 (\textcolor{blue}{+8.9}) & 33.2 (\textcolor{blue}{+11.2})\\
    \ding{52}&\ding{52} & \ding{52} & \ding{52} & & 40.2 (\textcolor{blue}{+9.1}) & 33.9 (\textcolor{blue}{+11.9})\\
    \ding{52}& & \ding{52} & \ding{52} & \ding{52} & 35.9 (\textcolor{blue}{+4.8})& 30.7 (\textcolor{blue}{+8.6})\\
    \ding{52}&\ding{52} & \ding{52} & \ding{52} & \ding{52} & \textbf{40.9} (\textcolor{blue}{+9.8}) & \textbf{34.1} (\textcolor{blue}{+12.1}) \\

    \bottomrule
    \end{tabular}
    \caption{\textbf{Ablation study of each component in our method.}}
    \label{tab:ablationcomponents}
\end{table}

\begin{table}[t]
    \centering
    \setlength{\tabcolsep}{6pt}
    \begin{tabular}{cccc|cc}
         \toprule
         Natural & Depth & Stylized & Edge & AR$_{100}^b$ & AR$_{100}^m$ \\
         \midrule
         \ding{52} & & & & 38.5 & 32.0 \\
         \ding{52} & \ding{52} & & & 40.5 & 33.3 \\
         \ding{52} &  & \ding{52} & & 40.2 & 33.5 \\
         \ding{52} & \ding{52} & & \ding{52} & 40.5 & 33.7 \\
         \ding{52} & \ding{52} & \ding{52} & & \textbf{40.9} &\textbf{34.1} \\
         \bottomrule
         
    \end{tabular}
    \caption{\textbf{Ablation study of different views used in our method.}}
    \label{tab:teston3views}
\end{table}

\noindent\textbf{Image transformation.}
We leverage colorized depth views with the help of additional auxiliary views to enforce the model to learn appearance-invariant representation.
To study the impact of the transformed views, we apply the off-the-shelf model to generate different transformed views on the COCO 2017 validation set.
We then study the impact of different views used and report the results in~\tabref{tab:teston3views}.
The model is trained on VOC classes and evaluated on Non-VOC classes.
When only one view is considered, we find that both depth maps and stylized images perform similarly.
By including an additional auxiliary view on top of the depth view, we observe a consistent improvement while adding stylized images perform slightly better than the edge map.

\begin{table}[t]
    \centering
    \setlength{\tabcolsep}{6pt}
    \begin{tabular}{l|ccc}
        \toprule
        AR$_{10}^{b}$ / AR$_{100}^{b}$ & Natural & Depth & Stylized \\
        \midrule
        CutLER~\cite{cutler} & 19.9 / 34.5 & 10.3 / 17.5 & 11.6 / 22.4 \\
        \emph{v}-CLR (\textbf{ours}) & \textbf{22.5} / \textbf{40.9} & \textbf{18.8} / \textbf{35.7} & \textbf{21.0} / \textbf{35.2} \\
        \bottomrule
    \end{tabular}
    \caption{\textbf{Evaluation results on three different views in the VOC$\rightarrow$Non-VOC setting.} Our method only uses natural images during inference, but it is also capable of processing multiple views.}
    \label{tab:vs_cutler}
\end{table}

\noindent\textbf{Comparison with CutLER.}
We leverage CutLER~\cite{cutler}, which possesses a satisfactory object-identifying ability, in our work to generate object proposals.
We compare the performance of CutLER as a detector versus our method on the novel Non-VOC classes in~\cref{tab:vs_cutler}.
While the performance margin is already 6.4\% in AR$_{100}^b$ between CutLER and our method on natural images, it is noticeable that the performance of CutLER degrades rapidly on these transformed images, evidenced by around 15\% performance gap on the two transformed views.
These results demonstrate that CutLER may suffer from potential textual bias, thus emphasizing the strength of learning appearance-invariant representation.

\begin{table}[t]
    \centering
    \setlength{\tabcolsep}{5pt}
    \begin{tabular}{l|ccc}
    \toprule
         & DINO-DETR~\cite{dino} & Ours & w/o Transform. \\
     \midrule
     AR$^b_{100}$ & 32.6 & \textbf{40.7} & 39.5 (\textcolor{blue}{-1.2})\\
     AR$^m_{100}$ & 26.9 & \textbf{33.8} & 32.4 (\textcolor{blue}{-1.4})\\
     \bottomrule
    \end{tabular}
    \caption{\textbf{Experiments in the VOC$\rightarrow$Non-VOC setting based on the Swin-Tiny~\cite{Liu2021SwinTH} backbone.}}
    \label{tab:swin-exp}
\end{table}

\noindent\textbf{Application to vision transformers.}
According to~\cite{Naseer2021IntriguingPO}, vision transformers exhibit less texture bias compared to CNNs. We thus additionally investigate the applicability of our method to vision transformers. We present experimental results utilizing the Swin-Tiny backbone in~\cref{tab:swin-exp}. Our approach significantly surpasses DINO-DETR~\cite{dino}, emphasizing the necessity of transformed views for enhanced performance with transformer-based architectures. These experimental results indicate that our method is applicable and can enhance the performance of vision transformer backbones.

\noindent\textbf{View choices and segmentation scenarios.}
We investigate the impact of incorporating various views on segmentation performance across different scenarios, including variations in object sizes and the number of instances.
Specifically, as shown in~\tabref{tab:numberins}, we analyze the effect of different combinations of views and evaluate the model under scenarios with varying instance counts.
The experimental results indicate that both depth maps and stylized images consistently improve performance across scenarios with diverse numbers of instances. 
Additionally, we assess the model's performance across different object sizes, as outlined in~\tabref{tab:layerins}, where objects are categorized into small, medium, and large based on the standard COCO dataset~\cite{coco}.
Our findings reveal that incorporating additional views significantly enhances performance on medium and large objects, while the improvements on small objects are relatively modest.

\begin{table}[!t]
    \centering
    \small
    \begin{subtable}[!htp]{0.5\textwidth}
        \centering
        \setlength{\tabcolsep}{3pt} 
        \begin{tabular}{l|ccccc}
        \toprule
     Count    & 1$\sim$3 & 4$\sim$6 & 7$\sim$9 &  $\geq$ 10 \\
         \midrule
      N & 58.8 / 48.7 & 42.4 / 35.8 & 33.5 / 28.1 & 21.6 / 17.5 \\
      N + D &  \textbf{60.9} / 50.3 & 44.4 / 37.1 & 35.4 / 29.5 & 23.3 / 18.7 \\
      N + S & 60.4 / 50.3 &  44.3 / 37.4 & 34.7 / 29.8 & 23.3 / 19.1 \\
      N + D + S &  60.6 / \textbf{50.5} & \textbf{44.7} / \textbf{37.6} & \textbf{35.9} / \textbf{30.3} & \textbf{24.0} / \textbf{19.9} \\
      \bottomrule 
        \end{tabular}
        \caption{Performance on scenarios with different number of instances} \label{tab:numberins}
    \end{subtable}%

    \begin{subtable}[!htp]{0.5\textwidth}
        \centering
        \small
        \setlength{\tabcolsep}{3pt} 
        \begin{tabular}{l|cccccc}
        \toprule
     Size    & Small & Medium & Large & All\\
         \midrule
      N &  16.6 / 12.3 & 45.3 / 38.1 & 73.9 / 63.5  & 38.5 / 32.0  \\
      N + D & 17.4 / 12.5 & 49.1 / 41.1 & 75.3 / 64.8 & 40.5 / 33.3  \\
      N + S & 17.1 / 12.7 & 48.8 / 41.1 & 75.2 / 65.0 & 40.2 / 33.5 \\
      N + D + S & \textbf{17.6} / \textbf{13.1} & \textbf{49.6} / \textbf{42.2} & \textbf{75.5} / \textbf{65.5} & \textbf{40.9} / \textbf{34.1} \\
      \bottomrule 
        \end{tabular}
        \caption{Performance on scenarios with different object sizes}\label{tab:layerins}
    \end{subtable}%
    \caption{\textbf{Ablation study of view choices on different segmentation scenarios in the VOC$\rightarrow$Non-VOC setting.} `N', `S', and `D' denote natural images, stylized images, and depth maps, respectively. We report AR$_{100}^b$ / AR$_{100}^m$ in the table.}\label{tab:segmentationscenarios}
\end{table}

\noindent\textbf{Detailed performance on unknown and known classes.}
To study the effect of our method on known and unknown classes, we train the model on a cross-dataset setting, VOC$\rightarrow$UVO, and evaluate the model on known and unknown classes, respectively.
As shown in~\tabref{tab:knownandunknown}, our method achieves performance comparable to the baseline model on known classes, while significantly improving recall on unknown objects by 10.7\% and across all classes by 8\% in terms of AR$^b_{100}$.
These results highlight the effectiveness of our method in discovering novel objects.

\begin{table}[!t]
    \centering
    \small
    \setlength{\tabcolsep}{3pt}
    \begin{tabular}{l|cc|cc|cc}
            \toprule
\multicolumn{1}{c|}{\multirow{2}{*}{Method}} & \multicolumn{2}{c|}{Known} & \multicolumn{2}{c|}{Unknown} & \multicolumn{2}{c}{All} \\
\multicolumn{1}{c|}{}                        & AR$_{100}^b$         & AR$_{100}^m$         & AR$_{100}^b$         & AR$_{100}^m$          & AR$_{100}^b$        & AR$_{100}^m$        \\ \midrule
DINO-DETR & 59.3  & \textbf{48.3} & 36.5 & 27.7 & 42.3 & 33.2 \\
\emph{v}-CLR (\textbf{ours}) & \textbf{60.9} & 47.0 & \textbf{47.2} & \textbf{35.9} & \textbf{50.3} & \textbf{38.4} \\ \bottomrule
    \end{tabular}
    \caption{\textbf{Evaluation results on known and unknown classes in the VOC$\rightarrow$UVO setting.}}\label{tab:knownandunknown} 
\end{table}

\begin{table}[!t]
    \centering
    \small 
    \setlength{\tabcolsep}{5pt}
    \begin{tabular}{l|ccc}
        \toprule
        Ratio & Small & Medium & Large \\ \midrule
        Ground-truth of Known Classes & 31.1\% & 34.9\% & 34.0\% \\
        + Proposals &  19.9\% & 28.5\% & 51.6\% \\ 
        \bottomrule
    \end{tabular}
    \caption{\textbf{Ratio of small, medium and large objects in the supervision.} The ratio is measured under COCO$\rightarrow$Objects365.}\label{tab:ratoofsize}
\end{table}

\noindent\textbf{Performance on small objects.}
As shown in~\tabref{tab:coco2objects365}, our method exhibits unstable performance on small objects.
Specifically, our method achieves an approximate 1\% improvement on small objects when applied to Deformable-DETR~\cite{deformabledetr}, but leads to performance degradation when applied to DINO-DETR~\cite{dino}. We attribute this instability to an imbalance in the ratio of objects with different sizes. In~\tabref{tab:ratoofsize}, we measure the size distribution of objects and observe that the ratio of small objects decreases significantly when incorporating proposals. This imbalance arises due to the CutLER~\cite{cutler} network's inherent preference for large objects, stemming from its pretraining process.

\section{Conclusion}\label{sec:conclusion}
To encourage the model to utilize appearance-invariant cues to discover objects, we propose a learning framework, named view-Consistent LeaRning (\emph{v}-CLR), for segmenting instances in an open world.
Specifically, our method randomly picks one from natural images, depth images, and an auxiliary view as input during training.
In this way, the model will tend to learn common features between the three views, which is beneficial for novel object discovery.
Besides, to help the model learn appearance-invariant features, we design a consistent objective based on the general object proposals.
The superiority of our approach is thoroughly validated with extensive experiments on cross-category and cross-dataset settings and consistently achieving state-of-the-art performance.

\noindent \textbf{Acknowledgement.} 
This work is supported by National Natural Science Foundation of China (Grant No. 62306251), Hong Kong Research Grant Council - Early Career Scheme (Grant No. 27208022), 
and HKU Seed Fund for Basic Research.
The computations were performed partly using research computing facilities offered by Information Technology Services, The University of Hong Kong.

{
    \small
    \bibliographystyle{ieeenat_fullname}
    \bibliography{ref}

\begin{thebibliography}{78}
\providecommand{\natexlab}[1]{#1}
\providecommand{\url}[1]{\texttt{#1}}
\expandafter\ifx\csname urlstyle\endcsname\relax
  \providecommand{\doi}[1]{doi: #1}\else
  \providecommand{\doi}{doi: \begingroup \urlstyle{rm}\Url}\fi

\bibitem[Ballester and Araujo(2016)]{ballester2016performance}
Pedro Ballester and Ricardo Araujo.
\newblock On the performance of googlenet and alexnet applied to sketches.
\newblock In \emph{AAAI}, 2016.

\bibitem[Bhat et~al.(2023)Bhat, Birkl, Wofk, Wonka, and M{\"u}ller]{bhat2023zoedepth}
Shariq~Farooq Bhat, Reiner Birkl, Diana Wofk, Peter Wonka, and Matthias M{\"u}ller.
\newblock Zoedepth: Zero-shot transfer by combining relative and metric depth.
\newblock \emph{arXiv preprint arXiv:2302.12288}, 2023.

\bibitem[Brendel and Bethge(2019)]{brendel2019approximating}
Wieland Brendel and Matthias Bethge.
\newblock Approximating cnns with bag-of-local-features models works surprisingly well on imagenet.
\newblock \emph{arXiv preprint arXiv:1904.00760}, 2019.

\bibitem[Cai and Vasconcelos(2018)]{cai2018cascade}
Zhaowei Cai and Nuno Vasconcelos.
\newblock Cascade r-cnn: Delving into high quality object detection.
\newblock In \emph{IEEE Conf. Comput. Vis. Pattern Recog.}, 2018.

\bibitem[Cai and Vasconcelos(2019)]{cai2019cascade}
Zhaowei Cai and Nuno Vasconcelos.
\newblock Cascade r-cnn: High quality object detection and instance segmentation.
\newblock \emph{IEEE Trans. Pattern Anal. Mach. Intell.}, 2019.

\bibitem[Cai et~al.(2023)Cai, Liu, Wang, Ge, Zhang, and Huang]{cai2023align}
Zhi Cai, Songtao Liu, Guodong Wang, Zheng Ge, Xiangyu Zhang, and Di Huang.
\newblock Align-detr: Improving detr with simple iou-aware bce loss.
\newblock \emph{arXiv preprint arXiv:2304.07527}, 2023.

\bibitem[Carion et~al.(2020)Carion, Massa, Synnaeve, Usunier, Kirillov, and Zagoruyko]{detr}
Nicolas Carion, Francisco Massa, Gabriel Synnaeve, Nicolas Usunier, Alexander Kirillov, and Sergey Zagoruyko.
\newblock End-to-end object detection with transformers.
\newblock In \emph{Eur. Conf. Comput. Vis.}, 2020.

\bibitem[Caron et~al.(2021)Caron, Touvron, Misra, J{\'e}gou, Mairal, Bojanowski, and Joulin]{dinov1}
Mathilde Caron, Hugo Touvron, Ishan Misra, Herv{\'e} J{\'e}gou, Julien Mairal, Piotr Bojanowski, and Armand Joulin.
\newblock Emerging properties in self-supervised vision transformers.
\newblock In \emph{Int. Conf. Comput. Vis.}, 2021.

\bibitem[Chaudhari et~al.(2019)Chaudhari, Choromanska, Soatto, LeCun, Baldassi, Borgs, Chayes, Sagun, and Zecchina]{chaudhari2019entropy}
Pratik Chaudhari, Anna Choromanska, Stefano Soatto, Yann LeCun, Carlo Baldassi, Christian Borgs, Jennifer Chayes, Levent Sagun, and Riccardo Zecchina.
\newblock Entropy-sgd: Biasing gradient descent into wide valleys.
\newblock \emph{Journal of Statistical Mechanics: Theory and Experiment}, 2019.

\bibitem[Chen et~al.(2022)Chen, Chen, Zeng, and Wang]{chen2022group}
Qiang Chen, Xiaokang Chen, Gang Zeng, and Jingdong Wang.
\newblock Group detr: Fast training convergence with decoupled one-to-many label assignment.
\newblock \emph{arXiv preprint arXiv:2207.13085}, 2022.

\bibitem[Chen and He(2021)]{simsiam}
Xinlei Chen and Kaiming He.
\newblock Exploring simple siamese representation learning.
\newblock In \emph{IEEE Conf. Comput. Vis. Pattern Recog.}, 2021.

\bibitem[Chen et~al.(2020)Chen, Fan, Girshick, and He]{mocov2}
Xinlei Chen, Haoqi Fan, Ross Girshick, and Kaiming He.
\newblock Improved baselines with momentum contrastive learning.
\newblock \emph{arXiv preprint arXiv:2003.04297}, 2020.

\bibitem[Chen et~al.(2023)Chen, Huang, Li, Teng, Wang, Shao, Loy, and Sheng]{siamesedetr}
Ze-Sen Chen, Gengshi Huang, Wei Li, Jianing Teng, Kun Wang, Jing Shao, Chen~Change Loy, and Lu Sheng.
\newblock Siamese detr.
\newblock In \emph{IEEE Conf. Comput. Vis. Pattern Recog.}, 2023.

\bibitem[Cheng et~al.(2021)Cheng, Schwing, and Kirillov]{maskformer}
Bowen Cheng, Alex Schwing, and Alexander Kirillov.
\newblock Per-pixel classification is not all you need for semantic segmentation.
\newblock In \emph{Adv. Neural Inform. Process. Syst.}, 2021.

\bibitem[Choi et~al.(2021)Choi, Jung, Yun, Kim, Kim, and Choo]{Choi2021RobustNetID}
Sungha Choi, Sanghun Jung, Huiwon Yun, Joanne~Taery Kim, Seungryong Kim, and Jaegul Choo.
\newblock Robustnet: Improving domain generalization in urban-scene segmentation via instance selective whitening.
\newblock In \emph{IEEE Conf. Comput. Vis. Pattern Recog.}, 2021.

\bibitem[Everingham et~al.(2010)Everingham, Van~Gool, Williams, Winn, and Zisserman]{voc}
Mark Everingham, Luc Van~Gool, Christopher~KI Williams, John Winn, and Andrew Zisserman.
\newblock The pascal visual object classes (voc) challenge.
\newblock \emph{Int. J. Comput. Vis.}, 2010.

\bibitem[Gao et~al.(2022)Gao, Wang, Han, and Guo]{gao2022adamixer}
Ziteng Gao, Limin Wang, Bing Han, and Sheng Guo.
\newblock Adamixer: A fast-converging query-based object detector.
\newblock In \emph{IEEE Conf. Comput. Vis. Pattern Recog.}, 2022.

\bibitem[Gatys et~al.(2017)Gatys, Ecker, and Bethge]{gatys2017texture}
Leon~A Gatys, Alexander~S Ecker, and Matthias Bethge.
\newblock Texture and art with deep neural networks.
\newblock \emph{Current opinion in neurobiology}, 2017.

\bibitem[Geiger et~al.(2013)Geiger, Lenz, Stiller, and Urtasun]{geiger2013vision}
Andreas Geiger, Philip Lenz, Christoph Stiller, and Raquel Urtasun.
\newblock Vision meets robotics: The kitti dataset.
\newblock \emph{The international journal of robotics research}, 2013.

\bibitem[Geirhos et~al.(2018)Geirhos, Rubisch, Michaelis, Bethge, Wichmann, and Brendel]{geirhos2018imagenet}
Robert Geirhos, Patricia Rubisch, Claudio Michaelis, Matthias Bethge, Felix~A Wichmann, and Wieland Brendel.
\newblock Imagenet-trained cnns are biased towards texture; increasing shape bias improves accuracy and robustness.
\newblock \emph{arXiv preprint arXiv:1811.12231}, 2018.

\bibitem[Ghiasi et~al.(2021)Ghiasi, Cui, Srinivas, Qian, Lin, Cubuk, Le, and Zoph]{copypaste}
Golnaz Ghiasi, Yin Cui, Aravind Srinivas, Rui Qian, Tsung-Yi Lin, Ekin~D Cubuk, Quoc~V Le, and Barret Zoph.
\newblock Simple copy-paste is a strong data augmentation method for instance segmentation.
\newblock In \emph{IEEE Conf. Comput. Vis. Pattern Recog.}, 2021.

\bibitem[Girshick(2015)]{fastrcnn}
Ross Girshick.
\newblock Fast r-cnn.
\newblock In \emph{Int. Conf. Comput. Vis.}, 2015.

\bibitem[Gu et~al.(2021)Gu, Lin, Kuo, and Cui]{gu2021open}
Xiuye Gu, Tsung-Yi Lin, Weicheng Kuo, and Yin Cui.
\newblock Open-vocabulary object detection via vision and language knowledge distillation.
\newblock \emph{arXiv preprint arXiv:2104.13921}, 2021.

\bibitem[Gupta et~al.(2019)Gupta, Dollar, and Girshick]{lvis}
Agrim Gupta, Piotr Dollar, and Ross Girshick.
\newblock Lvis: A dataset for large vocabulary instance segmentation.
\newblock In \emph{IEEE Conf. Comput. Vis. Pattern Recog.}, 2019.

\bibitem[He et~al.(2016)He, Zhang, Ren, and Sun]{he2016deep}
Kaiming He, Xiangyu Zhang, Shaoqing Ren, and Jian Sun.
\newblock Deep residual learning for image recognition.
\newblock In \emph{IEEE Conf. Comput. Vis. Pattern Recog.}, 2016.

\bibitem[He et~al.(2017)He, Gkioxari, Doll{\'a}r, and Girshick]{maskrcnn}
Kaiming He, Georgia Gkioxari, Piotr Doll{\'a}r, and Ross Girshick.
\newblock Mask r-cnn.
\newblock In \emph{Int. Conf. Comput. Vis.}, 2017.

\bibitem[He et~al.(2020)He, Fan, Wu, Xie, and Girshick]{mocov1}
Kaiming He, Haoqi Fan, Yuxin Wu, Saining Xie, and Ross Girshick.
\newblock Momentum contrast for unsupervised visual representation learning.
\newblock In \emph{IEEE Conf. Comput. Vis. Pattern Recog.}, 2020.

\bibitem[He et~al.(2022)He, Chen, Xie, Li, Doll{\'a}r, and Girshick]{mae}
Kaiming He, Xinlei Chen, Saining Xie, Yanghao Li, Piotr Doll{\'a}r, and Ross Girshick.
\newblock Masked autoencoders are scalable vision learners.
\newblock In \emph{IEEE Conf. Comput. Vis. Pattern Recog.}, 2022.

\bibitem[Hu et~al.(2024)Hu, Sun, Wang, and Yang]{hu2024dac}
Zhengdong Hu, Yifan Sun, Jingdong Wang, and Yi Yang.
\newblock Dac-detr: Divide the attention layers and conquer.
\newblock \emph{Adv. Neural Inform. Process. Syst.}, 2024.

\bibitem[Huang et~al.(2023{\natexlab{a}})Huang, Geiger, and Zhang]{good}
Haiwen Huang, Andreas Geiger, and Dan Zhang.
\newblock Good: Exploring geometric cues for detecting objects in an open world.
\newblock In \emph{Int. Conf. Learn. Represent.}, 2023{\natexlab{a}}.

\bibitem[Huang et~al.(2023{\natexlab{b}})Huang, Chen, Li, Li, Li, Song, Yan, and Xiong]{Huang2023StylePC}
Wei Huang, Chang~Wen Chen, Yong Li, Jiacheng Li, Cheng Li, Fenglong Song, Youliang Yan, and Zhiwei Xiong.
\newblock Style projected clustering for domain generalized semantic segmentation.
\newblock In \emph{IEEE Conf. Comput. Vis. Pattern Recog.}, 2023{\natexlab{b}}.

\bibitem[Jia et~al.(2023)Jia, Yuan, He, Wu, Yu, Lin, Sun, Zhang, and Hu]{jia2023detrs}
Ding Jia, Yuhui Yuan, Haodi He, Xiaopei Wu, Haojun Yu, Weihong Lin, Lei Sun, Chao Zhang, and Han Hu.
\newblock Detrs with hybrid matching.
\newblock In \emph{IEEE Conf. Comput. Vis. Pattern Recog.}, 2023.

\bibitem[Johnson et~al.(2017)Johnson, Hariharan, Van Der~Maaten, Fei-Fei, Lawrence~Zitnick, and Girshick]{clevr}
Justin Johnson, Bharath Hariharan, Laurens Van Der~Maaten, Li Fei-Fei, C Lawrence~Zitnick, and Ross Girshick.
\newblock Clevr: A diagnostic dataset for compositional language and elementary visual reasoning.
\newblock In \emph{IEEE Conf. Comput. Vis. Pattern Recog.}, 2017.

\bibitem[Kalluri et~al.(2023)Kalluri, Wang, Wang, Chandraker, Torresani, and Tran]{udos}
Tarun Kalluri, Weiyao Wang, Heng Wang, Manmohan Chandraker, Lorenzo Torresani, and Du Tran.
\newblock Open-world instance segmentation: Top-down learning with bottom-up supervision.
\newblock \emph{arXiv preprint arXiv:2303.05503}, 2023.

\bibitem[Keskar et~al.(2016)Keskar, Mudigere, Nocedal, Smelyanskiy, and Tang]{keskar2016large}
Nitish~Shirish Keskar, Dheevatsa Mudigere, Jorge Nocedal, Mikhail Smelyanskiy, and Ping Tak~Peter Tang.
\newblock On large-batch training for deep learning: Generalization gap and sharp minima.
\newblock \emph{arXiv preprint arXiv:1609.04836}, 2016.

\bibitem[Kim et~al.(2022)Kim, Lin, Angelova, Kweon, and Kuo]{oln}
Dahun Kim, Tsung-Yi Lin, Anelia Angelova, In~So Kweon, and Weicheng Kuo.
\newblock Learning open-world object proposals without learning to classify.
\newblock \emph{IEEE Robotics and Automation Letters}, 2022.

\bibitem[Kim et~al.(2023{\natexlab{a}})Kim, Angelova, and Kuo]{kim2023region}
Dahun Kim, Anelia Angelova, and Weicheng Kuo.
\newblock Region-aware pretraining for open-vocabulary object detection with vision transformers.
\newblock In \emph{IEEE Conf. Comput. Vis. Pattern Recog.}, 2023{\natexlab{a}}.

\bibitem[Kim and Byun(2020)]{kim2020learning}
Myeongjin Kim and Hyeran Byun.
\newblock Learning texture invariant representation for domain adaptation of semantic segmentation.
\newblock In \emph{IEEE Conf. Comput. Vis. Pattern Recog.}, 2020.

\bibitem[Kim et~al.(2023{\natexlab{b}})Kim, Kim, and Kim]{Kim2023TextureLD}
Sunghwan Kim, Dae-Hwan Kim, and Hoseong Kim.
\newblock Texture learning domain randomization for domain generalized segmentation.
\newblock In \emph{Int. Conf. Comput. Vis.}, 2023{\natexlab{b}}.

\bibitem[Kirillov et~al.(2023)Kirillov, Mintun, Ravi, Mao, Rolland, Gustafson, Xiao, Whitehead, Berg, Lo, Doll{\'a}r, and Girshick]{Kirillov2023SegmentA}
Alexander Kirillov, Eric Mintun, Nikhila Ravi, Hanzi Mao, Chloe Rolland, Laura Gustafson, Tete Xiao, Spencer Whitehead, Alexander~C. Berg, Wan-Yen Lo, Piotr Doll{\'a}r, and Ross~B. Girshick.
\newblock Segment anything.
\newblock In \emph{Int. Conf. Comput. Vis.}, 2023.

\bibitem[Kuhn(1955)]{kuhn1955hungarian}
Harold~W Kuhn.
\newblock The hungarian method for the assignment problem.
\newblock \emph{Naval research logistics quarterly}, 1955.

\bibitem[Lee et~al.(2022)Lee, Seong, Lee, and Kim]{Lee2022WildNetLD}
Suhyeon Lee, Hongje Seong, Seongwon Lee, and Euntai Kim.
\newblock Wildnet: Learning domain generalized semantic segmentation from the wild.
\newblock In \emph{IEEE Conf. Comput. Vis. Pattern Recog.}, 2022.

\bibitem[Li et~al.(2022)Li, Zhang, Liu, Guo, Ni, and Zhang]{li2022dn}
Feng Li, Hao Zhang, Shilong Liu, Jian Guo, Lionel~M Ni, and Lei Zhang.
\newblock Dn-detr: Accelerate detr training by introducing query denoising.
\newblock In \emph{IEEE Conf. Comput. Vis. Pattern Recog.}, 2022.

\bibitem[Li et~al.(2023)Li, Zhang, Xu, Liu, Zhang, Ni, and Shum]{maskdino}
Feng Li, Hao Zhang, Huaizhe Xu, Shilong Liu, Lei Zhang, Lionel~M Ni, and Heung-Yeung Shum.
\newblock Mask dino: Towards a unified transformer-based framework for object detection and segmentation.
\newblock In \emph{IEEE Conf. Comput. Vis. Pattern Recog.}, 2023.

\bibitem[Lin et~al.(2014)Lin, Maire, Belongie, Hays, Perona, Ramanan, Doll{\'a}r, and Zitnick]{coco}
Tsung-Yi Lin, Michael Maire, Serge Belongie, James Hays, Pietro Perona, Deva Ramanan, Piotr Doll{\'a}r, and C~Lawrence Zitnick.
\newblock Microsoft coco: Common objects in context.
\newblock In \emph{Eur. Conf. Comput. Vis.}, 2014.

\bibitem[Liu et~al.(2022)Liu, Li, Zhang, Yang, Qi, Su, Zhu, and Zhang]{liu2022dab}
Shilong Liu, Feng Li, Hao Zhang, Xiao Yang, Xianbiao Qi, Hang Su, Jun Zhu, and Lei Zhang.
\newblock Dab-detr: Dynamic anchor boxes are better queries for detr.
\newblock \emph{arXiv preprint arXiv:2201.12329}, 2022.

\bibitem[Liu et~al.(2023)Liu, Ren, Chen, Zeng, Zhang, Li, Li, Huang, Su, Zhu, et~al.]{liu2023detection}
Shilong Liu, Tianhe Ren, Jiayu Chen, Zhaoyang Zeng, Hao Zhang, Feng Li, Hongyang Li, Jun Huang, Hang Su, Jun Zhu, et~al.
\newblock Detection transformer with stable matching.
\newblock \emph{arXiv preprint arXiv:2304.04742}, 2023.

\bibitem[Liu et~al.(2017)Liu, Cheng, Hu, Wang, and Bai]{liu2017richer}
Yun Liu, Ming-Ming Cheng, Xiaowei Hu, Kai Wang, and Xiang Bai.
\newblock Richer convolutional features for edge detection.
\newblock In \emph{IEEE Conf. Comput. Vis. Pattern Recog.}, 2017.

\bibitem[Liu et~al.(2021)Liu, Lin, Cao, Hu, Wei, Zhang, Lin, and Guo]{Liu2021SwinTH}
Ze Liu, Yutong Lin, Yue Cao, Han Hu, Yixuan Wei, Zheng Zhang, Stephen Lin, and Baining Guo.
\newblock Swin transformer: Hierarchical vision transformer using shifted windows.
\newblock In \emph{Int. Conf. Comput. Vis.}, 2021.

\bibitem[Meng et~al.(2021)Meng, Chen, Fan, Zeng, Li, Yuan, Sun, and Wang]{meng2021conditional}
Depu Meng, Xiaokang Chen, Zejia Fan, Gang Zeng, Houqiang Li, Yuhui Yuan, Lei Sun, and Jingdong Wang.
\newblock Conditional detr for fast training convergence.
\newblock In \emph{Int. Conf. Comput. Vis.}, 2021.

\bibitem[Naseer et~al.(2021)Naseer, Ranasinghe, Khan, Hayat, Khan, and Yang]{Naseer2021IntriguingPO}
Muzammal Naseer, Kanchana Ranasinghe, Salman~Hameed Khan, Munawar Hayat, Fahad~Shahbaz Khan, and Ming-Hsuan Yang.
\newblock Intriguing properties of vision transformers.
\newblock In \emph{Adv. Neural Inform. Process. Syst.}, 2021.

\bibitem[Oquab et~al.(2023)Oquab, Darcet, Moutakanni, Vo, Szafraniec, Khalidov, Fernandez, Haziza, Massa, El-Nouby, et~al.]{oquab2023dinov2}
Maxime Oquab, Timoth{\'e}e Darcet, Th{\'e}o Moutakanni, Huy Vo, Marc Szafraniec, Vasil Khalidov, Pierre Fernandez, Daniel Haziza, Francisco Massa, Alaaeldin El-Nouby, et~al.
\newblock Dinov2: Learning robust visual features without supervision.
\newblock \emph{arXiv preprint arXiv:2304.07193}, 2023.

\bibitem[Pont-Tuset et~al.(2016)Pont-Tuset, Arbelaez, Barron, Marques, and Malik]{mcg}
Jordi Pont-Tuset, Pablo Arbelaez, Jonathan~T Barron, Ferran Marques, and Jitendra Malik.
\newblock Multiscale combinatorial grouping for image segmentation and object proposal generation.
\newblock \emph{IEEE Trans. Pattern Anal. Mach. Intell.}, 2016.

\bibitem[Saito et~al.(2022)Saito, Hu, Darrell, and Saenko]{ldet}
Kuniaki Saito, Ping Hu, Trevor Darrell, and Kate Saenko.
\newblock Learning to detect every thing in an open world.
\newblock In \emph{Eur. Conf. Comput. Vis.}, 2022.

\bibitem[Saleh and Elgammal(2015)]{wikiart}
Babak Saleh and Ahmed Elgammal.
\newblock Large-scale classification of fine-art paintings: Learning the right metric on the right feature.
\newblock \emph{arXiv preprint arXiv:1505.00855}, 2015.

\bibitem[Shao et~al.(2019)Shao, Li, Zhang, Peng, Yu, Zhang, Li, and Sun]{objects365}
Shuai Shao, Zeming Li, Tianyuan Zhang, Chao Peng, Gang Yu, Xiangyu Zhang, Jing Li, and Jian Sun.
\newblock Objects365: A large-scale, high-quality dataset for object detection.
\newblock In \emph{Int. Conf. Comput. Vis.}, 2019.

\bibitem[Silberman et~al.(2012)Silberman, Hoiem, Kohli, and Fergus]{silberman2012indoor}
Nathan Silberman, Derek Hoiem, Pushmeet Kohli, and Rob Fergus.
\newblock Indoor segmentation and support inference from rgbd images.
\newblock In \emph{Eur. Conf. Comput. Vis.}, 2012.

\bibitem[Teng et~al.(2023)Teng, Liu, Guo, and Wang]{teng2023stageinteractor}
Yao Teng, Haisong Liu, Sheng Guo, and Limin Wang.
\newblock Stageinteractor: Query-based object detector with cross-stage interaction.
\newblock \emph{arXiv preprint arXiv:2304.04978}, 2023.

\bibitem[Wang et~al.(2024)Wang, Vaze, and Han]{wang2024dissect}
Hongjun Wang, Sagar Vaze, and Kai Han.
\newblock Dissecting out-of-distribution detection and open-set recognition: A critical analysis of methods and benchmarks.
\newblock \emph{Int. J. Comput. Vis.}, 2024.

\bibitem[Wang et~al.(2021{\natexlab{a}})Wang, Feiszli, Wang, and Tran]{uvo}
Weiyao Wang, Matt Feiszli, Heng Wang, and Du Tran.
\newblock Unidentified video objects: A benchmark for dense, open-world segmentation.
\newblock In \emph{Int. Conf. Comput. Vis.}, 2021{\natexlab{a}}.

\bibitem[Wang et~al.(2022{\natexlab{a}})Wang, Feiszli, Wang, Malik, and Tran]{ggn}
Weiyao Wang, Matt Feiszli, Heng Wang, Jitendra Malik, and Du Tran.
\newblock Open-world instance segmentation: Exploiting pseudo ground truth from learned pairwise affinity.
\newblock In \emph{IEEE Conf. Comput. Vis. Pattern Recog.}, 2022{\natexlab{a}}.

\bibitem[Wang et~al.(2021{\natexlab{b}})Wang, Zhang, Shen, Kong, and Li]{wang2021dense}
Xinlong Wang, Rufeng Zhang, Chunhua Shen, Tao Kong, and Lei Li.
\newblock Dense contrastive learning for self-supervised visual pre-training.
\newblock In \emph{IEEE Conf. Comput. Vis. Pattern Recog.}, 2021{\natexlab{b}}.

\bibitem[Wang et~al.(2022{\natexlab{b}})Wang, Yu, De~Mello, Kautz, Anandkumar, Shen, and Alvarez]{wang2022freesolo}
Xinlong Wang, Zhiding Yu, Shalini De~Mello, Jan Kautz, Anima Anandkumar, Chunhua Shen, and Jose~M Alvarez.
\newblock Freesolo: Learning to segment objects without annotations.
\newblock In \emph{IEEE Conf. Comput. Vis. Pattern Recog.}, 2022{\natexlab{b}}.

\bibitem[Wang et~al.(2023{\natexlab{a}})Wang, Girdhar, Yu, and Misra]{cutler}
Xudong Wang, Rohit Girdhar, Stella~X Yu, and Ishan Misra.
\newblock Cut and learn for unsupervised object detection and instance segmentation.
\newblock In \emph{IEEE Conf. Comput. Vis. Pattern Recog.}, 2023{\natexlab{a}}.

\bibitem[Wang et~al.(2023{\natexlab{b}})Wang, Shen, Yuan, Du, Li, Hu, Crowley, and Vaufreydaz]{wang2023tokencut}
Yangtao Wang, Xi Shen, Yuan Yuan, Yuming Du, Maomao Li, Shell~Xu Hu, James~L Crowley, and Dominique Vaufreydaz.
\newblock Tokencut: Segmenting objects in images and videos with self-supervised transformer and normalized cut.
\newblock \emph{IEEE Trans. Pattern Anal. Mach. Intell.}, 2023{\natexlab{b}}.

\bibitem[Wilms et~al.(2024)Wilms, Rolff, Hillemann, Johanson, and Frintrop]{SOS}
Christian Wilms, Tim Rolff, Maris Hillemann, Robert Johanson, and Simone Frintrop.
\newblock Sos: Segment object system for open-world instance segmentation with object priors.
\newblock In \emph{Eur. Conf. Comput. Vis.}, 2024.

\bibitem[Wu and Deng(2022)]{Wu2022SingleDomainGO}
Aming Wu and Cheng Deng.
\newblock Single-domain generalized object detection in urban scene via cyclic-disentangled self-distillation.
\newblock In \emph{IEEE Conf. Comput. Vis. Pattern Recog.}, 2022.

\bibitem[Wu et~al.(2023)Wu, Jiang, Yan, Lu, Yuan, and Luo]{sword}
Jiannan Wu, Yi Jiang, Bin Yan, Huchuan Lu, Zehuan Yuan, and Ping Luo.
\newblock Exploring transformers for open-world instance segmentation.
\newblock In \emph{Int. Conf. Comput. Vis.}, 2023.

\bibitem[Wu et~al.(2021)Wu, Hu, Sheng, and Xu]{wu2021styleformer}
Xiaolei Wu, Zhihao Hu, Lu Sheng, and Dong Xu.
\newblock Styleformer: Real-time arbitrary style transfer via parametric style composition.
\newblock In \emph{Int. Conf. Comput. Vis.}, 2021.

\bibitem[Xie and Tu(2015)]{Xie2015HolisticallyNestedED}
Saining Xie and Zhuowen Tu.
\newblock Holistically-nested edge detection.
\newblock \emph{Int. J. Comput. Vis.}, 2015.

\bibitem[Yang and Soatto(2020)]{yang2020fda}
Yanchao Yang and Stefano Soatto.
\newblock Fda: Fourier domain adaptation for semantic segmentation.
\newblock In \emph{IEEE Conf. Comput. Vis. Pattern Recog.}, 2020.

\bibitem[Zhang et~al.(2021)Zhang, Bengio, Hardt, Recht, and Vinyals]{zhang2021understanding}
Chiyuan Zhang, Samy Bengio, Moritz Hardt, Benjamin Recht, and Oriol Vinyals.
\newblock Understanding deep learning (still) requires rethinking generalization.
\newblock \emph{Communications of the ACM}, 2021.

\bibitem[Zhang et~al.(2024)Zhang, Zhong, and Han]{zhang2024mr}
Chang-Bin Zhang, Yujie Zhong, and Kai Han.
\newblock Mr. detr: Instructive multi-route training for detection transformers.
\newblock \emph{arXiv preprint arXiv:2412.10028}, 2024.

\bibitem[Zhang et~al.(2022)Zhang, Li, Liu, Zhang, Su, Zhu, Ni, and Shum]{dino}
Hao Zhang, Feng Li, Shilong Liu, Lei Zhang, Hang Su, Jun Zhu, Lionel~M Ni, and Heung-Yeung Shum.
\newblock Dino: Detr with improved denoising anchor boxes for end-to-end object detection.
\newblock \emph{arXiv preprint arXiv:2203.03605}, 2022.

\bibitem[Zhang et~al.(2025)Zhang, Zhang, Cui, Sun, Cao, Guo, Han, Zhao, Wang, Sun, et~al.]{zhang2025humanoidpano}
Qiang Zhang, Zhang Zhang, Wei Cui, Jingkai Sun, Jiahang Cao, Yijie Guo, Gang Han, Wen Zhao, Jiaxu Wang, Chenghao Sun, et~al.
\newblock Humanoidpano: Hybrid spherical panoramic-lidar cross-modal perception for humanoid robots.
\newblock \emph{arXiv preprint arXiv:2503.09010}, 2025.

\bibitem[Zhang et~al.(2018)Zhang, Xiang, Hospedales, and Lu]{zhang2018deep}
Ying Zhang, Tao Xiang, Timothy~M Hospedales, and Huchuan Lu.
\newblock Deep mutual learning.
\newblock In \emph{IEEE Conf. Comput. Vis. Pattern Recog.}, 2018.

\bibitem[Zhu et~al.(2023)Zhu, Li, Chen, Fan, Mao, Jing, Liu, and Shen]{segprompt}
Muzhi Zhu, Hengtao Li, Hao Chen, Chengxiang Fan, Weian Mao, Chenchen Jing, Yifan Liu, and Chunhua Shen.
\newblock Segprompt: Boosting open-world segmentation via category-level prompt learning.
\newblock In \emph{Int. Conf. Comput. Vis.}, 2023.

\bibitem[Zhu et~al.(2020)Zhu, Su, Lu, Li, Wang, and Dai]{deformabledetr}
Xizhou Zhu, Weijie Su, Lewei Lu, Bin Li, Xiaogang Wang, and Jifeng Dai.
\newblock Deformable detr: Deformable transformers for end-to-end object detection.
\newblock \emph{arXiv preprint arXiv:2010.04159}, 2020.

\end{thebibliography}
}

\clearpage
\section*{Appendix}
\appendix
\section{Experimental Details}
\subsection{Auxiliary Views}

\begin{figure*}[!htp] 
  \centering
  \small
  \begin{overpic}[width=.99\linewidth]{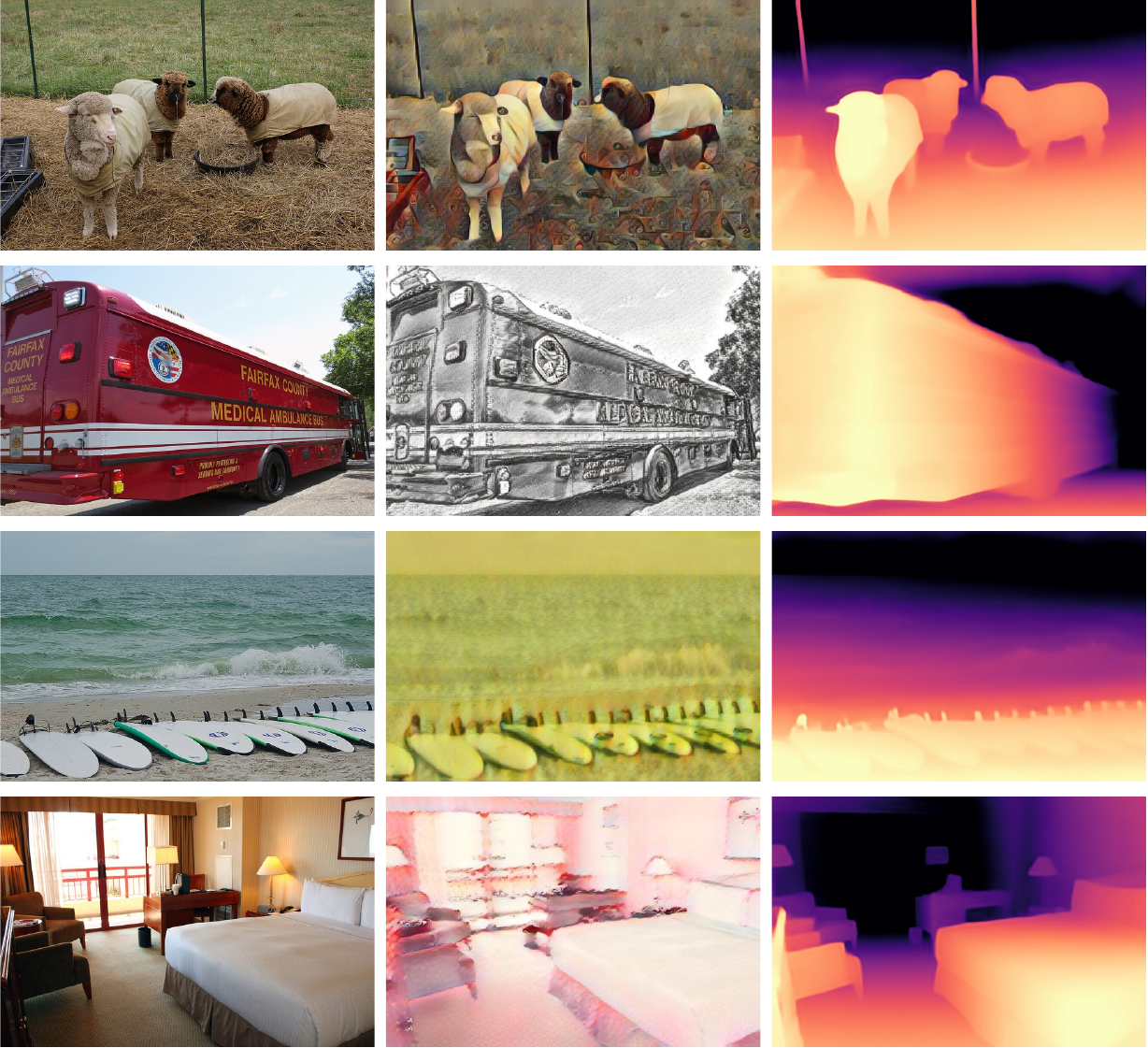} 
  \end{overpic}
  \vskip 0.1in
  \caption{\textbf{Visualization of three views} used in our method, natural, art-stylized, and colorized depth images, respectively.
  }\label{fig:views}
\end{figure*}

Our learning framework leverages multiple transformed views of the original natural image. Specifically, we apply off-the-shelf models to transform natural images into art-stylized and colorized depth images. For the art-stylized transformation, we utilize the pre-trained StyleFormer~\cite{wu2021styleformer} model, which is trained on the WikiArt~\cite{wikiart} dataset. For each natural image, we randomly select a target style from the WikiArt~\cite{wikiart} dataset. For the colorized depth transformation, we employ the off-the-shelf ZoeDepth~\cite{bhat2023zoedepth} model, pre-trained on the NYU Depth v2~\cite{silberman2012indoor} and KITTI~\cite{geiger2013vision} datasets. 
Additionally, for edge maps used in our ablation study, we apply the off-the-shelf RCF~\cite{liu2017richer} model for edge detection. Examples of natural, art-stylized, and colorized depth images are shown in~\figref{fig:views}. Notably, no human annotations are used for generating depth maps or stylized images, ensuring that our method avoids any information leakage.

\subsection{Experiments on the CLEVR Dataset}

The CLEVR dataset~\cite{clevr} is a synthetic dataset featuring objects characterized by four attributes:
\begin{itemize}
    \item Size: large, small
    \item Shape: cube, sphere, cylinder
    \item Color: gray, red, blue, green, brown, purple, cyan, yellow
    \item Material: rubber, metal
\end{itemize}
In this work, we focus on two attributes—color and material—for illustrative simplicity. Specifically, we designate \emph{red metal} objects as the known class, while objects with any other attribute combination are treated as unknown classes. The CLEVR dataset~\cite{clevr} comprises 70,000 training images and 15,000 validation images. We apply vanilla DINO-DETR~\cite{dino} and train the model under two settings: with and without colorized depth images and stylized images. When using colorized depth images, the model randomly selects either a natural image, a depth map or a stylized image as input, each with equal probability. With 300 denoising queries in DINO-DETR~\cite{dino}, we train the model for 2,000 iterations with a batch size of 8, while retaining the remaining training configurations identical to those of vanilla DINO-DETR~\cite{dino}.

\subsection{Object Proposal Generation}

Thanks to large-scale self-supervised learning, neural networks have shown remarkable capabilities in object recognition and localization~\cite{wang2023tokencut, oquab2023dinov2}.
Leveraging this advancement, unsupervised instance segmentation~\cite{cutler, wang2022freesolo} has recently achieved significant progress.
By benefiting from unsupervised training, these methods exhibit strong instance awareness, making them well-suited for generating object proposals in our work.
Throughout this paper, we employ the ImageNet-pretrained Cascade R-CNN~\cite{cai2019cascade} from CutLER~\cite{cutler} to infer object proposals from the dataset.
For each training image, we apply Non-Maximum Suppression (NMS) with a threshold of 0.7 and select the top-10 proposals based on prediction confidence.

\begin{figure}[!t]
    \centering
    \includegraphics[width=0.7\linewidth]{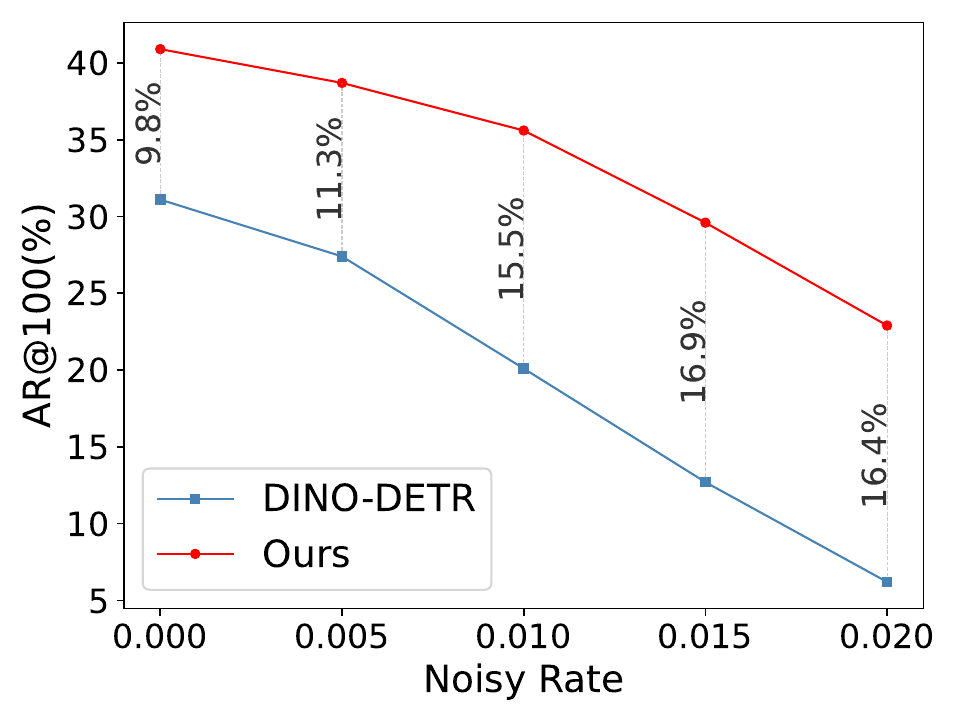}
    \caption{\textbf{AR$_{100}^b$ under different noisy rates.} All models are evaluated in the VOC$\rightarrow$Non-VOC setting.}
    \label{fig:perturbation}
\end{figure}

\begin{algorithm}[t]
\caption{Pseudo-code of Parameter Perturbation in a PyTorch-like style.}
\label{alg:noise}
\definecolor{codeblue}{rgb}{0.25,0.5,0.5}
\lstset{
  backgroundcolor=\color{white},
  basicstyle=\fontsize{9pt}{9pt}\ttfamily\selectfont,
  columns=fullflexible,
  breaklines=true,
  captionpos=b,
  commentstyle=\fontsize{9pt}{9pt}\color{codeblue},
  keywordstyle=\fontsize{9pt}{9pt},
}
\scriptsize
\begin{lstlisting}[language=python]
# image: input image tensors
# model: the detector
# noise_std: the standard deviation of gaussian noise
def perturbation_forward(image, model, noise_std):
    # adding gaussian noise for each parameter
    for name, param in model.named_parameters():
        param += torch.randn_like(param) * noise_std
    output = model(image)    
    return output
\end{lstlisting}
\end{algorithm}

\begin{figure*}[htp] 
  \centering
  \small
  \begin{overpic}[width=1.0\linewidth]{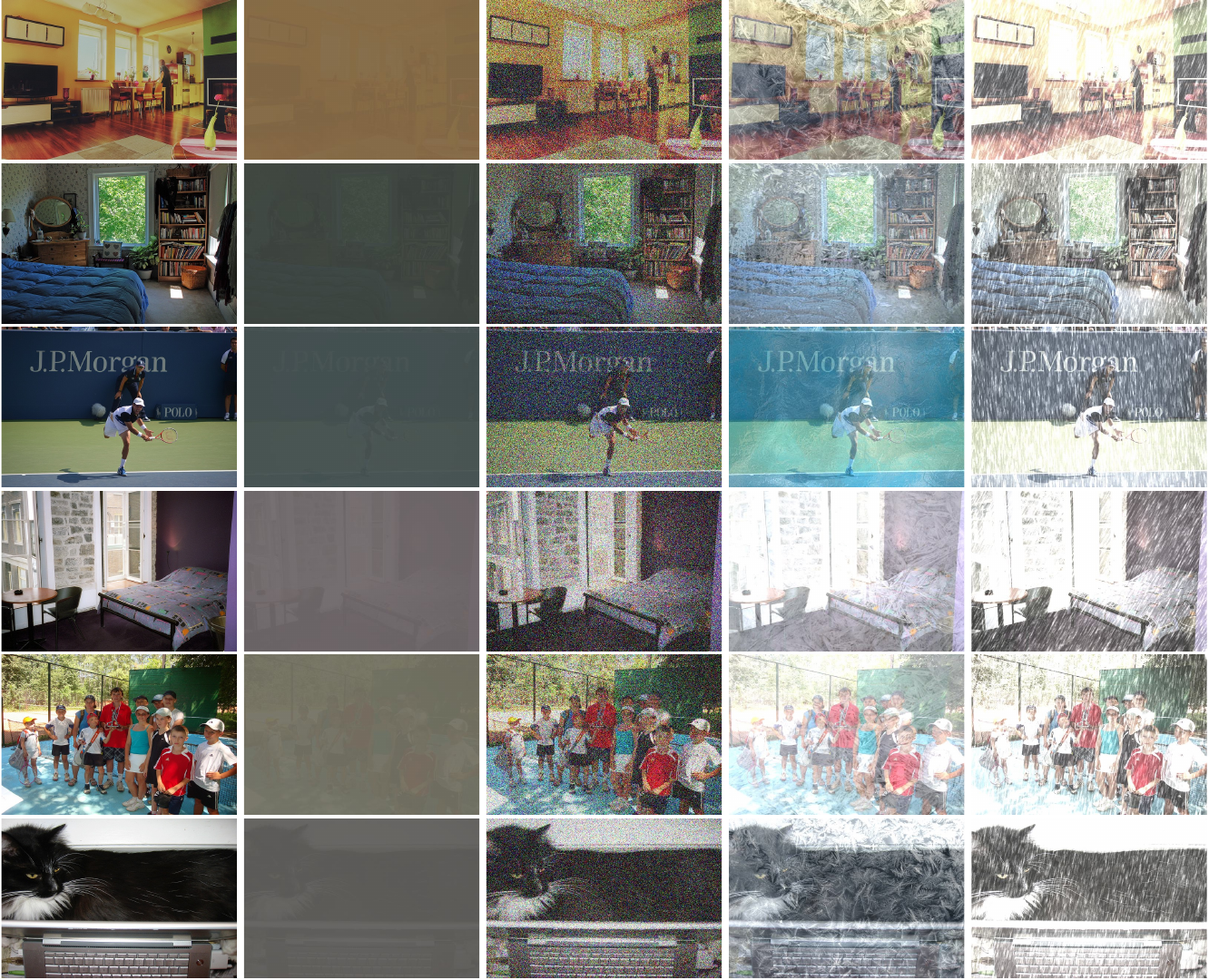} 
  \put(8, -2){Natural}
  \put(27, -2){Contrast}
  \put(44, -2){Gaussian Noise}
  \put(68, -2){Frost}
  \put(88, -2){Snow}
  \end{overpic}
  \vskip 0.2in
  \caption{\textbf{Examples of distorted images on COCO 2017~\cite{coco} validation set.}
  }\label{fig:distortions}
\end{figure*}

\section{Robustness against Parameter Perturbation}
Numerous studies~\cite{chaudhari2019entropy, zhang2021understanding, keskar2016large, zhang2018deep} have demonstrated that neural networks trained with flatten minima exhibit superior generalization ability, \ie, the minima of the model should be in wide valleys rather than narrow crevices~\cite{chaudhari2019entropy,zhang2021understanding,keskar2016large,zhang2018deep}.
In such cases, small perturbations to model parameters should not significantly degrade the performance of a model with strong generalization ability.
Consequently, we can assess a model’s generalization by introducing random perturbations to its parameters. Specifically, as detailed in Alg.~\ref{alg:noise}, we inject Gaussian noise with varying standard deviations into all network parameters and evaluate the resulting performance.
All models are trained on VOC classes and evaluated on Non-VOC classes. As illustrated in Fig.~\ref{fig:perturbation}, we define the noise rate as the standard deviation of the Gaussian noise. With increasing noise rates, both our model and DINO-DETR experience performance degradation. However, at high noise rates, our method consistently outperforms the baseline by a substantial margin, demonstrating greater robustness to parameter perturbations.

\begin{figure*}[t]
  \centering
  \begin{subfigure}[b]{0.24\textwidth}
    \includegraphics[width=\textwidth]{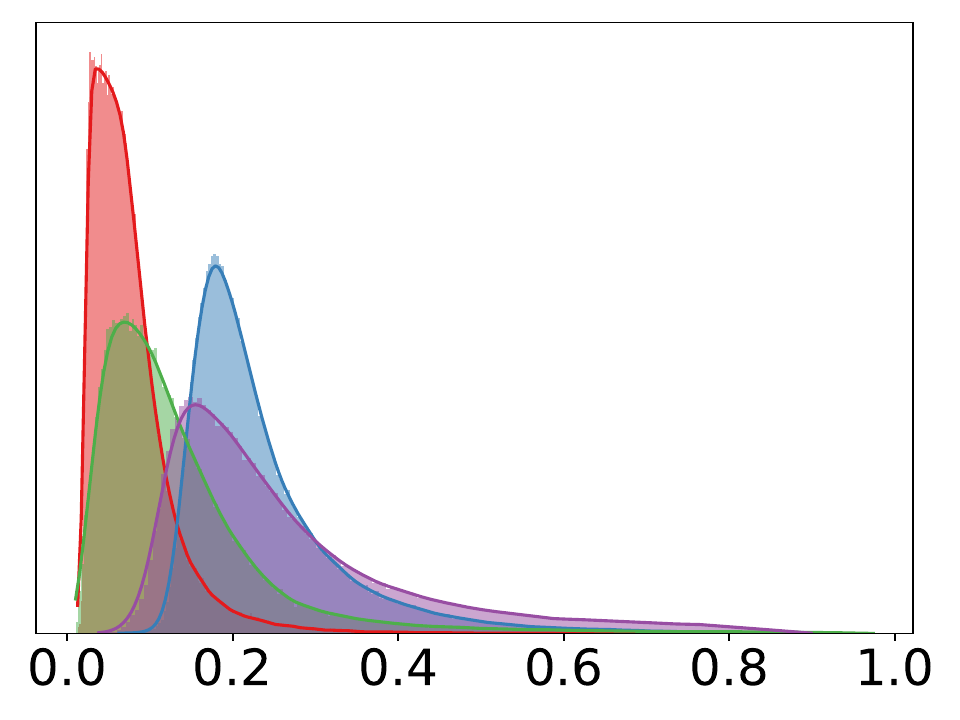}
    \caption{Contrast}
    \label{fig:subfigure_a}
  \end{subfigure}
  \begin{subfigure}[b]{0.24\textwidth}
    \includegraphics[width=\textwidth]{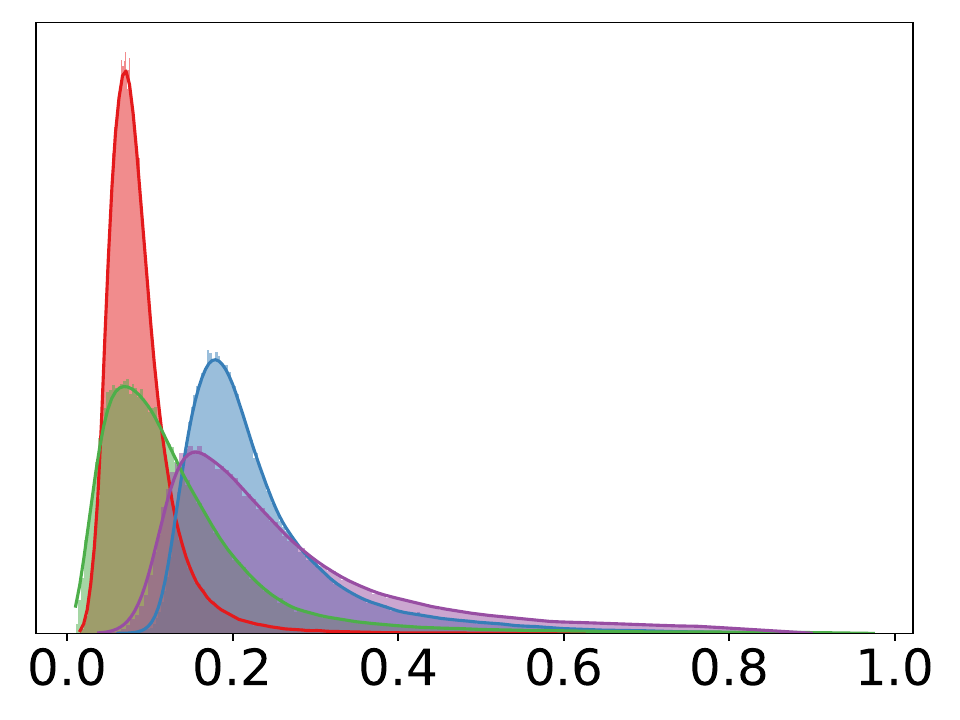}
    \caption{Gaussian Noise}
    \label{fig:subfigure_b}
  \end{subfigure}
  \begin{subfigure}[b]{0.24\textwidth}
    \includegraphics[width=\textwidth]{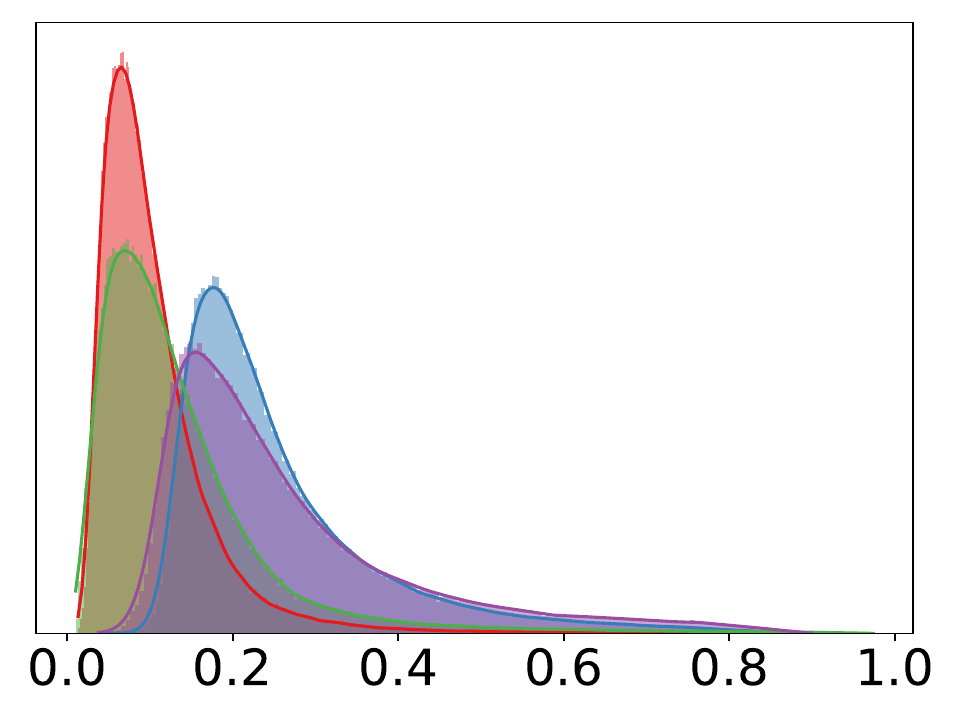}
    \caption{Snow}
    \label{fig:subfigure_c}
  \end{subfigure}
  \begin{subfigure}[b]{0.24\textwidth}
    \includegraphics[width=\textwidth]{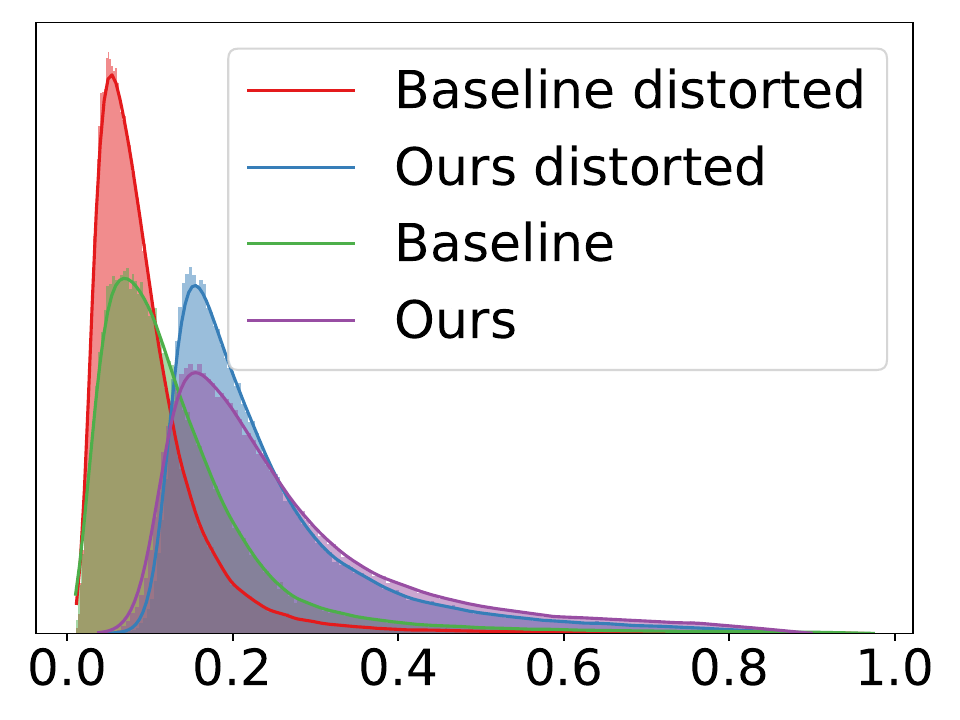}
    \caption{Frost}
    \label{fig:subfigure_d}
  \end{subfigure}
  \vskip 0.1in
  \caption{\textbf{The distribution of prediction scores from the baseline DINO-DETR~\cite{dino} and our \emph{v}-CLR under four types of image distortion.} For visualization clarity, we calculate the distribution of \emph{top}-50 prediction scores.}
  \label{fig:dist-score}
\end{figure*}

\section{Robustness against Image Distortion}
To verify the effectiveness of our method under different input perturbations, we evaluate our model under four popular distortions,  Contrast, Gaussian Noise, Snow, and Frost.
As shown in~\figref{fig:distortions}, we generate validation images with these distortions and evaluate the model’s performance on them. 
We examine the robustness of our \emph{v}-CLR approach against different types of image distortions.
In Fig.~\ref{fig:dist-score}, we plot the distribution of prediction scores for both the baseline DINO-DETR~\cite{dino} and our method, with and without image distortions. Our method (purple) consistently yields higher prediction scores than the baseline (green) on undistorted images. For distorted images, the distribution of the distorted baseline (red) exhibits a heavier right tail compared to the undistorted baseline (green), indicating that distortions reduce DINO-DETR’s prediction confidence. In contrast, our method demonstrates greater robustness to image distortions, as the distributions of prediction scores for distorted and undistorted images show similar right-tail behavior. Surprisingly, the prediction score distribution for our method on distorted images exhibits even lower variance and a slightly higher mean than on undistorted images. This further suggests that image distortions have minimal impact on our model’s prediction confidence.

\section{Comparison with SiameseDETR}
We further compare our method with Siamese DETR~\cite{siamesedetr}, a recent self-supervised DETR-like object detector. Siamese DETR employs two augmented views to enforce instance-level consistency. Although it also utilizes transformations, its motivation differs substantially from ours, and the transformations in Siamese DETR do not specifically address texture bias. We evaluate both methods in the Non-VOC$\rightarrow$VOC setting, as shown in Tab.~\ref{tab:vs_siamesedetr}, ensuring a fair comparison since VOC classes are unknown to both models. Experimental results reveal that our method surpasses Siamese DETR by a significant margin across all evaluation metrics, underscoring the effectiveness of our proposed framework.

\begin{table}[t]
    \centering
    \setlength{\tabcolsep}{11pt}
    \begin{tabular}{l|ccc}
    \toprule
    Method & AR$_1^b$ & AR$_{10}^b$ & AR$_{100}^{b}$ \\
    \midrule
    SiameseDETR~\cite{siamesedetr} & 12.4 & 23.0 & 30.7 \\
    \emph{v}-CLR (\textbf{ours}) & \textbf{16.8} & \textbf{42.7} & \textbf{60.2} \\
    \bottomrule
    \end{tabular}
    \caption{\textbf{Comparison with Siamese DETR~\cite{siamesedetr}.} For a fair comparison, all experiments are conducted on the Non-VOC$\rightarrow$VOC setting with Deformable-DETR~\cite{deformabledetr}.}
    \label{tab:vs_siamesedetr}
\end{table}

\end{document}